\documentclass[12pt, oneside]{article}   	% use "amsart" instead of "article" for AMSLaTeX format
\usepackage{geometry}                		% See geometry.pdf to learn the layout options. There are lots.
\usepackage{graphicx}				% Use pdf, png, jpg, or epsÂ§ with pdflatex; use eps in DVI mode
\usepackage{amssymb}
\usepackage{amsmath}
\usepackage{stmaryrd}
\usepackage{natbib}
\usepackage{hyperref}
\usepackage{linguex}
\usepackage{tikz}
\usepackage{nicefrac}
\usepackage{xcolor}
\usepackage{booktabs}
\usepackage{diagbox}
\newcommand{\sem}[1]{\ensuremath{\llbracket #1 \rrbracket}}

\usepackage[toc,page]{appendix}

\usepackage[all]{nowidow}

\hypersetup{colorlinks=true, citecolor=blue}

\newcommand{\pe}[1]{\textcolor{black}{#1}}
\newcommand{\bs}[1]{\textcolor{black}{#1}}

\title{Explaining vague language\thanks{Acknowledgements: We are most grateful to Joel Sobel for very detailed comments and suggestions on the first draft of this paper, and to two anonymous referees for helpful reviews. We also thank Tansu Alpcan, Kenny Chatain, Alexandre Cremers, Ad\`ele Henot-Mortier, Christina Pawlowitsch, David Spector, Steven Verheyen, and audiences in Adelaide, Canberra, Melbourne, and Paris, for helpful exchanges on the topic of this paper. PE gratefully acknowledges Monash University (Philosophy Department) and Melbourne University (EEE Department) for their hospitality and support during the course of this research. This work was supported by PLEXUS (Grant Agreement no 101086295), a Marie Sk\l odowska-Curie action funded by the EU under the Horizon Europe Research and Innovation Programme.
}}
\author{Paul \'Egr\'e\thanks{paul.egre@cnrs.fr, IRL Crossing, CNRS / Department of Philosophy, ENS-PSL.} \and Benjamin Spector\thanks{benjamin.spector@ens.psl.eu, Institut Jean-Nicod, CNRS, ENS-PSL, EHESS. Department of Cognitive Studies of ENS-PSL.}}
\date{}%June 13, 2025}							% Activate to display a given date or no date

\begin{document}
\maketitle

\begin{abstract} \noindent Why is language vague? Vagueness may be explained and rationalized if it can be shown that vague language is more useful to speaker and hearer than precise language. In a well-known paper, Lipman proposes a game-theoretic account of vagueness in terms of mixed strategy that leads to a puzzle: vagueness cannot be strictly better than precision at equilibrium. More recently, \'Egr\'e, Spector, Mortier and Verheyen have put forward a Bayesian account of vagueness establishing that using vague words can be strictly more informative than using precise words. This paper proposes to compare both results and to explain why they are not in contradiction. Lipman's definition of vagueness relies exclusively on a property of signaling strategies, without making any assumptions about the lexicon, whereas \'Egr\'e et al.'s involves a layer of semantic content. We argue that the semantic account of vagueness is needed, and more adequate and explanatory of vagueness. %\pe{Besides, while our account puts a limitation on the expressiveness of the language, we show that the existence of a semantically vague message is necessary for speakers to convey information about the shape of their probability distribution.} 

\end{abstract}

\newpage

\section{Introduction}

Vagueness is one of the puzzling features of natural languages. Whereas some words, like ``even'' and ``odd'' as applied to numbers, have a clear mathematical definition with determinate boundaries of application, others like ``large'' and ``small'' lack such a definition and do not appear to have a determinate extension. Instead, they admit borderline cases, cases that are uncertain and that leave room for variability both between and within speakers, and they also give rise to well-known paradoxes (soritical reasoning, see \citealt{kamp1981paradox}). How come that so many words and expressions of natural language are like ``large'' and ``small'', and unlike ``even'' and ``odd''? Why doesn't a word like ``large'', applied to discrete quantities, invariably mean ``larger than 50'', or ``larger than 100''? 

This question has been of interest to philosophers, linguists, and economists, since at least the 1900s (see \citealt{borel1907economic,egre2014borel}). Various answers have been proposed. One line of response is that vagueness results from our limited powers of discrimination, implying that if we have words like ``small'' and ``large'', it is due to our imperfect ability to distinguish between close quantities, even in the discrete domain (see \citealt{williamson1994vagueness,luce1956semiorders,franke2018vagueness}). %According to that view, vague language is not the right explanandum, instead the possibility of a precise language is what should be puzzling us. 
Another proposal is that vagueness may originate in the way in which we acquire our concepts: we do not appear to learn the meaning of ``large'' and ``small' by means of a rigid rule or definition, but instead, by reference to specific objects or exemplars, which can be grouped in different and open-ended ways (see \citealt{waismann1945verifiability,wittgenstein1953philosophical}). Yet a third response is that there is an inherent link between vagueness and context-sensitivity: ``large'' cannot invariably mean ``larger than 50'', because what counts as large depends on a comparison class, and also on expectations that may vary depending on the speaker and listener (see \citealt{sapir1944grading}).

These various responses offer promising and interconnected answers to the problem raised. In this paper, we are interested in the game-theoretical approach originally proposed by \citealt{lipman2009why}, in what has become a classic paper \pe{(published in this special issue)}. Roughly presented, the leading idea is that vagueness could be explained and rationalized if it can be shown that the vague use of a word is more useful or beneficial to speaker and hearer, in a situation of cooperative exchange, than the precise use of that word. Lipman, however, presents in his paper a set of negative results, challenging the reader to find a better response.
Lipman's results state that there is no scenario in which a vague language is strictly better than a precise language at equilibrium. Recently, however, \cite{egre2023optimality} have put forward a Bayesian account of vague communication that appears to directly contradict this result. What their paper shows is that there exist scenarios in which vague language is strictly better than precise language. This tension is puzzling, and our first goal in this paper is to explain and to resolve it. 

The results are not in contradiction, fundamentally because they do not use the same definition of vagueness. Lipman's main definition involves the notion of a mixed strategy over signals. It thereby corresponds to a purely pragmatic account of the notion of meaning, on which the meaning of an expression, relative to a speaker's strategy, corresponds to the set of states that are mapped to this signal by the speaker's strategy with nonzero probability. By contrast, Egr\'e et al.'s definition of vagueness is semantic: it implies that an expression is vague provided it has \textit{truth-conditions} involving an open parameter. On that account, the speaker's strategy operates on a level of structure that is missing from the purely pragmatic approach, and the difference between vague and precise expressions is not merely dependent on the speaker's use (or strategy), but on that level of truth-conditional content. 

Our second goal, based on that clarification, is to draw broader lessons regarding game-theoretic approaches of meaning: focusing on the case of vagueness, we may view Lipman's negative results as a reductio of a bare pragmatic account of meaning, in which messages are not pre-loaded with content, and as reaffirming the need for a Gricean division of labor between semantics and pragmatics. Our broader goal, finally, is to shed light on the nature of linguistic vagueness, in particular by showing the importance of a definition in terms of open truth conditions. %\pe{Considerations about soritical reasoning will be left aside, however, since they do not pertain directly to the main issue we discuss.} 

In section \ref{sec:lipman}, we start with a review of \cite{lipman2009why}'s results, with a \pe{critical} examination of his definition of vagueness. In section \ref{sec:account}, we present the optimality result of \cite{egre2023optimality}, and in particular the main assumption it makes regarding the way in which semantic vagueness is recruited to communicate the shape of one's epistemic uncertainty. \pe{We indicate how this property, originally stated for the case of a particular vague word, may be assumed to hold for any vague expression. %{fine-grained epistemic states}. %we explain why it does not contradict the former.
%In section \ref{sec:generalizing}, we examine how the strategy used to show the optimality of a specific vague expression could be generalized to arbitrary vague expressions. 
As presented there, however, these results are established }\bs{in a very simple setting where the speaker is assumed to talk to a literal listener} \pe{who uses Bayesian update but only to interpret messages literally.}
%who is not a Bayesian agent but simply interprets messages literally.  
\bs{In section \ref{sec:game} we provide a a proper game-theoretic setting in which a speaker chooses a message under the assumption that the listener \pe{is a Bayesian agent who} knows the speaker's strategy \pe{and can reason about it}. % and is a Bayesian agent. 
We define a specific equilibrium concept, the Iterated Best Response equilibrium (based on \citealt{franke2009signal,franke2011quantity} and \citealt{goodman2013knowledge}). This leads to a stronger result: in our setup, in the absence of messages whose conventional meaning directly refers to the speaker's probability distribution, the existence of a semantically vague message is necessary in order for a speaker to be able communicate information about the \emph{shape} or their probability distribution (and not just its \emph{support}) at equilibrium.} %

In section \ref{sec:discussion}, finally, we explain why our account does not logically contradict Lipman's results, and we draw more general lessons regarding the semantics-pragmatics division for linguistic meaning. \pe{We also discuss the interaction between the semantic characteristic of vague expressions at the heart of our account, and our assumption that the language's expressiveness is limited.} We \pe{defer the proof of the main result to an appendix}, since one of the goals of this paper is to provide a conceptual and philosophical clarification on the assumptions and implications of both models that is easily accessible to non-experts.

%On that approach, the meaning of an expression $e$ is defined in purely pragmatic terms relative to a strategy $s$, as the set of states that 
%It thereby corresponds to a purely pragmatic account of the notion of meaning, on which the meaning of an expression can be identified with the class of situations that the 

%Lipman in his paper proposes a definition of vagueness which implies some negative results, challenging the reader to find a better response. In a nutshell, Lipman's 

%a word like ``tall'' instead of only words like ``taller than 6 feet'', or ``taller than the median"? The problem has been of interest to game-theorists, in particular to B. Lipman in an influential paper. Lipman's paper is of the aporetical genre: Lipman formulates a negative result, and challenges the reader with finding a better response. 

\section{Suboptimal Vagueness}\label{sec:lipman}

In this section we provide an informal review and a critical discussion of Lipman's characterization of vagueness, and of its suboptimality results. Lipman's main definition involves the notion of mixed strategy over signals. In passing, he provides a weakening of that definition using a relevance criterion. We explain why both definitions fall short of adequately characterizing vagueness.

\subsection{Vagueness as mixed strategy}\label{sec:mixed}

%How does Lipman characterize a vague language? 
Let us consider Lipman's main ingredients in his account of vagueness.
First there is a \textit{cardinality} constraint on the language. According to Lipman, %\BS{for a language to be vague,}
{for vagueness to be potentially useful,} there needs to be fewer messages in the language than there are things to describe.
%\\\BS{NB: I had a look at Lipman: this is not a definition, Lipman just says that when that's not the case it"'s obvious that mixed strategies are necessarily suboptima}.

 %According to Lipman, \BS{or a language to be vague, there needs to be fewer messages in the language than there are things to describe.{I had a look at Lipman : this is not a definition, Lipman just says that when that's not the case it"'s obvious that mixed strategies are necessarily suboptimal. I reorganized the bedining of this section}\\*
 This condition actually echoes \cite{russell1923vagueness}'s definition of vagueness, according to which ``a representation is vague when the relation of the representing system to the represented system is not one-one,
but one-many.'' But as stressed by Lipman, and already by Russell, this is not a sufficient condition for vagueness, since for example a language may have two messages ``taller than 185cm'', and ``shorter than 185cm", and the language would be coarse-grained but precise, it would draw a clear-cut partition of the set of heights, assumed to be infinite, or even just finite but larger than the set of messages. Still it is a necessary condition, since otherwise, with as many words as there are heights, the language could achieve a bijection between representing symbols and represented things, and would be thereby precise.

Secondly, Lipman introduces a \textit{pragmatic} condition on language use. Lipman argues that if a speaker were to always pair observations with messages in a constant way, then the language would not be vague. For the language to be vague, according to Lipman, there must be at least one observation $o$ such that on some occasions the speaker uses one message $m$ to describe it, and on some other occasions a different message $m'$. Lipman therefore links vagueness to the use by the speaker of a mixed strategy, namely a probabilistic function from observations to messages. Lipman's definition of vagueness in that regard inverts the one proposed by Russell, for it implies that the relation of the \textit{represented} system to the \textit{representing system} is one-many. For a pure strategy, the relation of an observation to a signal is constant, each signal is mapped to exactly one expression. When using a mixed strategy, however, the relation of an observation to a signal becomes variable: the same observation can produce different signals. 

%For example, a pure strategy would be to say always ``tall'' when observing a height of 180cm, and another pure strategy to say always ``short'' when observing a height of 180cm. A nondegenerate mixed strategy based on those two is one in which the speaker uses ``tall'' on some occasions with a positive probability, and ``short'' on others with a positive probability, so that the same observation receives a probabilistic message from the speaker. For a listener, whose goal is to infer what observation the speaker had in mind, this can create uncertainty about what the speaker observed. 

To use one of Lipman's examples, assume there are three heights which are $180, 185, 190$, and two messages $short$ and $ tall$. Then a pure strategy for the speaker is just a functional mapping from heights to signals, in which, to every height, there corresponds a unique signal. For example, a function mapping $180$ and $185$ to $short$, and $190$ to $tall$ is a pure speaker strategy. A nondegenerate mixed strategy, on the other hand, is one in which for at least one observation, the speaker can use the two signals $short$ and $tall$ with positive probability for each. For example, it may be a strategy where $180$ is mapped to $short$, $190$ to $tall$, and $185$ to either with probability $1/2$. A non-degenerate mixed strategy is a non-functional mapping this time, since the same observation can be described by different messages.

Another way of presenting the difference behind Lipman's identification of vagueness with a mixed strategy is in terms of the properties of the speaker's strategy $s$ viewed as a mapping from the set $O$ of observations to the set $M$ of messages (expressions, or signals). Given a strategy $s$, and $m\in M$, let $s^{-1}(m)$ be the set of observations in $O$ that are mapped to $m$ with positive probability, and let us call $s^{-1}(m)$ the \textit{speaker's meaning} for $m$. The set of $s^{-1}(m)$ for all $m\in M$ is a \textit{partition} of $O$ when $s$ is a pure strategy, but when $s$ is not pure it is no longer a partition, but a \textit{cover} of $O$, whose cells can overlap. In the above example, for instance, the pure strategy produces the partition $\{\{180,185\}, \{190\}\}$, whereas the mixed strategy produces the cover $\{\{180,185\}, \{185,190\}\}$. From the listener's perspective, therefore, the latter situation introduces an uncertainty about the speaker's meaning which is absent in the pure strategy case. Despite its plausibility, however, below we will argue that Lipman's characterization of vagueness is insufficient. To see this, we need to look at Lipman's suboptimality results.

%Whereas the use of a pure strategy makes the meaning of some symbol {\it general} given the structural condition (it applies to more than one observation), the use of a mixed strategy makes the meaning of some symbol not just general but also {\it variable}. 

%When using a pure strategy, the relation of an observation to a signal is constant, each signal is mapped to exactly one expression. But conversely, by the structural condition on the language, the relation of a signal to an observation is variable: the same signal can apply to several observations. When using a mixed strategy, however, the relation of an observation to a signal is no longer constant, it is also variable: the same observation can produce different signals. Lipman's definition of vagueness in that regard inverts the one proposed by Russell, for basically it says that the relation between of the \textit{represented} system to the \textit{representing system} is one-many. 

%So Lipman's idea is that in a precise language, the relation between an observation and a signal is constant for the speaker, but in a vague language it can be variable.

\subsection{Lipman's suboptimality results}\label{sec:suboptimal}

To rationalize vagueness, one must explain how vague language can be more beneficial to speakers of a language community than precise language. The assumption is that of a cooperative game-theoretic setting, in which the speaker's utility and hearer's utility are aligned. In that setting, using Lipman's identification of vague language with the use of a mixed strategy, there must then be a Nash equilibrium involving a nondegenerate mixed strategy by the speaker over messages, which outperforms Nash equilibria based on using pure strategies.

Lipman's puzzle about vagueness is that this can't happen: for any mixed strategy Nash Equilibrium, there is is a pure strategy Nash Equilibrium that does equally well. The result rests on a fundamental property of mixed strategies, which is that each of the pure strategies in the support of a mixed strategy Nash equilibrium must itself be a best response to the other player's strategy, and therefore yield the same expected payoff. Indeed, if one of those pure strategies gave a higher payoff, then the player would have an interest to deviate and just use that strategy. And if one of those strategies gave a worse payoff, then the mixed strategy could also be made better by never playing that pure strategy, and so by not including it in the support of the mixed strategy.

Based on this negative result, Lipman entertains a modification of the definition of vagueness. \pe{We can think of it in relation to \textit{questions}. %, viewed as partitions on the set of observations. 
The listener is interested in knowing whether a proposition $p$ holds, and the question whether $p$ can be represented as a partition between the states in which $p$ is true, and the states in which $p$ is false (see \citealt{hamblin1958questions,groenendijk1982semantic}).} Thus, Lipman calls a language \textit{precise} when it is attached to a pure strategy in equilibrium which maps observations that are in the same equivalence class vis a vis a question to the same message, assuming that observations that are in the same equivalence class induce the same preferences over actions. A language is called \textit{vague} otherwise. According to that definition, if at equilibrium the speaker uses a message that only modifies the listener's prior distribution, without settling the question even partially, that is without allowing the listener to rule out at least one cell of the partition, then the message is considered vague. %So vague in that sense appears to mean: unable to settle the question, even partially.

As an example of such vagueness, Lipman considers three states of the world, $h_1,h_2,h_3$, and two messages $m$ and $m'$. The assumption is that in equilibrium, the speaker uses $m$ to signal $\{h_1,h_3\}$, and $m'$ to signal $\{h_2\}$. If the listener is interested in knowing whether $\{h_3\}$ or $\{h_1,h_2\}$, and assuming her prior is 1/3 on each height, then the listener's posterior on $\{h_1,h_2\}$ after hearing the message $m$ is 1/2 when her prior was initially 2/3, but that's not enough to settle the question. The message is called ``vague'' by Lipman here, in the sense that the observations $h_1$, $h_2$ are connected to different messages, despite these two heights answering the same question.

%\clearpage

Lipman's second suboptimality result is that if there is a Nash equilibrium in the signaling game, then there is a pure strategy Nash equilibrium whose associated language for the speaker is precise, basically by the construction of a strategy that ensures that observations will be locked in to the same message for all members of the same equivalence class. In the previous example, such a strategy could be constructed by mapping $h_1,h_2$ to the same message and $h_3$ to another, so by allowing messages to basically line up with the partition corresponding to the question.

Let us take stock. According to Lipman's first definition of vagueness, a language is vague when it involves a mixed strategy over signals, in which messages do not indicate the speaker's meaning categorically to the listener. According to Lipman's second definition of vagueness, there are two ways in which a language can be vague: either because the language involves a mixed strategy over signals, as per the first definition, or because it does not allow the listener to partially settle a question of interest, even when it relies on a pure strategy over signals. Lipman's results show that irrespective of which definition is favored, vagueness cannot be optimal.

Both of Lipman's results are puzzling, since when we look at specific vague words, the nonoptimality theorems appear to run against casual observations. Consider vague words like \textit{tall} and \textit{short} in relation to the first result: we do appear to use them in ways that are fluctuating. In particular, we agree that borderline cases between \textit{tall} and \textit{short} may indeed be viewed as cases to which a speaker would apply \textit{tall} some of the time, and \textit{short} some of the time (see \citealt{mccloskey1978natural,douven2013vagueness,egre2013vagueness}). And consider the same words \textit{tall} and \textit{short} in relation to the second result: when asked ``how tall is this person?'', it does appear informative to hear ``John is tall'' in response to the question, even if that answer may not allow one to resolve that John is above any precise height (see \citealt{lassiter2017adjectival}). So what could be missing in Lipman's approach?

\subsection{Is vagueness adequately characterized?}
\label{sec:objections}

First of all, we can set aside the definition of vagueness that makes a message vague when it is irrelevant with regard to a question. For if in relation to the question ``was Dana born on a Monday or on a Tuesday?'' the answer given were ``Dana is 28 years old", and the speaker and listener knew originally that Dana could be either 28 or 48, and born either on a Monday or on a Tuesday, we are in a case of an irrelevant answer. In this case, however, none of the properties mentioned is vague in the relevant sense. While the use of vague words in answers to questions can \textit{coincide} with the property of irrelevance described by Lipman, this feature is not a defining property of vagueness.

So the real issue concerns Lipman's identification of vagueness with the existence of a mixed strategy.
{We think this identification is inadequate, even though within the framework Lipman assumes, it might be the only reasonable option.   %As mentioned, In Lipman's approach, meaning is fully identified with use-conditions. 
Vagueness is traditionally characterized by the existence of so-called \textit{borderline cases}, \pe{also known as \textit{dubious} or \textit{uncertain} cases of application of a predicate}  (\citealt{borel1907economic,black1937vagueness,williamson1994vagueness,smith2008vagueness}).\footnote{\pe{The existence of borderline cases is generally agreed to be at least a necessary condition for vagueness. Whether it is sufficient is moot. It has been argued that a predicate could not be vague without also being sorites-susceptible (see \citealt{fine1975vagueness,bueno2012vagueness}). In this paper, we leave further considerations about the sorites paradox aside, however, as they are tangential to the main issue, also in Lipman's paper. Nevertheless, all of the predicates we model and consider in our discussion have borderline cases and are prone to soritical reasoning.}} For instance, even if I know precisely Gloria's height, I might be unsure whether she counts as tall, because her height falls in this grey zone of heights where we are not quite sure whether \textit{tall} applies - Gloria would then be a borderline case of tallness. In Lipman's framework, in which meaning is identified with use conditions, the only conceivable way to approximate this characterization is to say that there are situations where one may or may not use the word \textit{tall}, i.e. making the choice of messages probabilistic. Considering the pair \textit{short} and \textit{tall}, the fact that there are cases that might be borderline cases of both tallness and shortness (\citealt{egrezehr2018}) translates, in this framework, into there being states in which the speaker's strategy is a non-degenerate mixed strategy over the pair $(\textrm{\textit{short}}, \textrm{\textit{tall}})$.}

{However, we view this characterization as flawed, in that it fails to capture our intuitive, pre-theoretical concept of vagueness, or, for that matter, the concept of vagueness as it has been characterized in philosophical logic, philosophy of language and formal semantics (see \citealt{sorensen2023vagueness} for an overall introduction). Among other things, Lipman's definition implies that certain theoretically possible patterns of uses involve vagueness, while they would not in fact fit either the intuitive notion or the corresponding one espoused by philosophers and linguists. Furthermore, and most importantly, Lipman's definition has the unfortunate consequence that there cannot be any overlap in meaning between two messages without such messages being vague.} 

 Consider a language with two words \textit{rhombus}, and \textit{rectangle}, and suppose the domain contains three kinds of geometrical figures: rectangles that are not squares, squares, and then rhombi that are not squares. Here is a possible mixed strategy for the speaker: utter \textit{rhombus} for a rhombus that is not a square with probability 1, utter \textit{rectangle} for a rectangle that is not a square with probability 1, and then utter \textit{rhombus} half the time and \textit{rectangle} half the time when it is a square (see Figure \ref{fig:rectangle}).

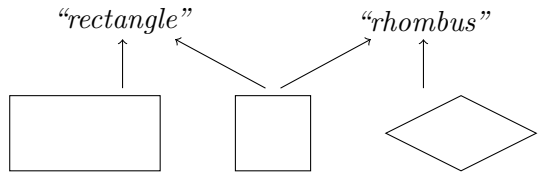
\begin{figure}
  \centering
  \begin{tikzpicture}
    % Rectangle
    \draw (0,0) rectangle (2,1);
    \node at (1.5,2) {\textit{``rectangle''}}; % Text label inside the rectangle
    % Arrow pointing to "rectangle"
    \draw[->] (1.5,1.1) -- (1.5,1.75);
    
    % Square
    \draw (3,0) rectangle (4,1);
    
    % Rhombus
    \draw (6,0) -- (7,0.5) -- (6,1) -- (5,0.5) -- cycle;
     \node at (5.5,2) {\textit{``rhombus''}}; % Text label inside the rectangle
    
    % Arrow pointing to "rhombus"
    \draw[->] (5.5,1.1) -- (5.5,1.75);
    
    \draw[->] (3.4,1.1) -- (2.2,1.75);

\draw[->] (3.6,1.1) -- (4.8,1.75);

  \end{tikzpicture}
  \caption{Rectangle, Square, and Rhombus}
\end{figure}\label{fig:rectangle}

According to Lipman, this would make the meanings of ``rhombus'' and ``rectangle'' vague. But one may object that in this case, the concepts of rectangle and rhombus simply overlap on squares, {and that the words are used in a way that is consistent with their \textit{precise} mathematical meanings in English}. %\BS{}{Even if one arbitrarily picks between \textit{rhombus } and \textit{rectangle} to refer to a square, such a use would not thereby make a square a borderline case of a rectangle or of a rhombus, nor the meaning of the words vague.} 
{For sure, the strategy we described might not be the best possible one, and in real life people will use the word \textit{square} to refer to squares, and approximate thereby a pure strategy. However, given that  the words \textit{rectangle} and \textit{rhombus} have overlapping denotations in English, this must mean, in Lipman's framework, that English speakers use a mixed a strategy when they want to refer to a square, since the only notion of meaning Lipman countenances identifies meaning with use. Yet a square is not a borderline-case of a rectangle, nor  a borderline-case of a rhombus, it is a rather a clear case of both. The general point is the following:  Lipman's approach does not allow for a notion of meaning which would not \textit{ipso facto} characterize two messages whose denotations overlap as being vague, since in this framework overlapping denotations can only be characterized as overlap in contexts of use.}  \bs{Importantly, we do not dispute Lipman's result according to which a mixed strategy is never superior to every pure strategy. The point of our example is purely conceptual: according to Lipman's definition, the mixed strategy we describe (completely independently of the question whether it could be a speaker's strategy at equilibrium, or whether it is, in some precise sense, an optimal strategy) would make ``rhombus'' and ``rectangle'' vague by definition. We argue that this is a flawed definition, which does not capture one of the defining properties of vagueness, namely the existence of borderline-cases.}

There is a potential way out for Lipman. Recall that a speaker's strategy is a function from states to probability distributions over messages. One might argue that states are very fine-grained, so that even though \textit{rectangle} and \textit{rhombus} can be both used to refer to squares, \pe{they} are in fact never used in exactly the same states, so that English speakers are in fact using a pure strategy when talking about geometrical figures, and the meaning of \textit{square} and \textit{rectangle} would be defined directly in terms of the states in which we use them, and not as referring to geometric figures. We don't think this is a satisfactory answer.  Imagine that a language contains two exact synonyms $X$ and $Y$ and that a speaker randomly chooses between $X$ and $Y$ when referring to something that falls under their denotation, with absolutely no influence of the fine-grained aspects of the context of utterance. Whether such a situation exists is not relevant to our argument; our point is simply that in such a hypothetical situation, this would not provide any ground whatsoever to characterize $X$ and $Y$ as vague. Imagine for instance a mathematician who would randomly select between the two equivalent statements ``$x \leq y$'' and ``$y \geq x$''. It is very clear that the mathematician's use of symbols would not in any sense make these two statements vague\bs{, according to both the intuitive sense of the word ``vague'' and existing characterizations in philosophy.}

Lipman may respond that the use of a mixed strategy over signals is \pe{only a} necessary condition for the language to be vague. \pe{Hence, mixed strategies over synonyms would not thereby instantiate vagueness.} {As discussed below, we will not even grant that \pe{the use of a mixed strategy is necessary to characterize vagueness}, but we acknowledge that this might seem reasonable to a first approximation}. The reason is that a borderline case of a vague predicate like ``tall'', for example, can indeed be characterized as one for which a competent speaker is uncertain {as to whether the word applies, and so might be such}  that if the speaker has to choose between ``tall'' and its negation ``not tall'', the speaker may use both on different occasions, or even on the same occasion (\citealt{ripley2011contradictions,alxatib2011psychology}).%\BS{This view of borderline status is more constrained than the view according to which the same case may be given different names, but it involves the notion of negation, and in that regard a level of meaning that is absent from the consideration of just use conditions.}{}

More generally, we consider that Lipman's approach basically fails to distinguish between what we might call ``making \textit{vague use} of a word'', and ``making use of a \textit{vague word}''. Differently put, the main limitation of Lipman's approach is the lack of an incorporation of semantic vagueness in his model. On his account, signals have no internal structure, and vagueness is only a matter of the variable use of words relative to an observation. As stressed by several theorists of vagueness, however, vagueness is not just a matter of use, it is a matter of meaning. For example, an adequate account of vagueness needs to explain what semantically distinguishes the vague adjective ``tall'' from the precise adjectival expression ``taller than 180cm'', or what makes the expressionq ``{noon}'' and ``{between 11.56pm and 12.04pm}'' precise and the expression ``noonish'' vague.

Before discussing the notion of semantic vagueness, we want to emphasize another aspect of Lipman's suboptimality results. Lipman in his discussion of vagueness entertains a further possibility to rationalize vagueness, connected to the notion of speaker uncertainty. Lipman writes: 

\begin{quote}
 ``Perhaps the most obvious argument for the usefulness of vague terms is that one cannot observe the height of an individual precisely enough to be sure in all cases of how to classify someone in such a precise language.''

\end{quote}

We agree with this observation, in fact we think it is one of the ingredients of the rationalization of vague language. The underlying idea, put forward by several authors, is that by using ``this person is tall'', a speaker with imperfect discrimination commits less than by uttering ``this person is taller than 180cm'' (see \citealt{russell1923vagueness, channell1994vague, krifka2007approximate}). However, Lipman considers that this line of explanation too will fall prey to his main suboptimality result. His response is that an uncertain speaker will form a probability distribution about heights, but that in agreement with his theorem, the use of a mixed strategy to signal this probability distribution will remain suboptimal compared to the use of a pure strategy.

%\BS{There is, however, a loophole here, if we can show that using a pure strategy mapping probability distributions to observations is in fact compatible with the use of vague words being \emph{more informative} than precise messages in specific cases. In the next section, we turn to giving such an account.}
{We do not disagree, but we will not ourselves identify vagueness with mixed strategies, and we will argue that using vague messages sometimes allows speakers to be maximally informative about their epistemic states --- with the result that a perfectly rational speaker might rely on a pure strategy mapping epistemic states to messages, some of which will be vague.}{}

% In the next section, we will show that the consideration of this level of semantic structure gives us resources to explain the utility of vagueness.

\section{Optimal Vagueness}\label{sec:account}

In this section we present the approach proposed in \cite{egre2023optimality} regarding the way in which using a vague word can present an informational benefit to speaker and listener, compared to using a precise expression. \bs{We focus on a very simple model of communication where the listener is simply conditioning her prior probability distribution on the information contained in the literal meaning of a message.\footnote{\bs{\citeauthor{egre2023optimality}'s \citeyear{egre2023optimality}~proposal is couched within the Rational Speech Act model  (\citealt{goodman2013knowledge}), which defines a sequence of speakers and listeners of increasing sophistication, and is related to our proposal in section \ref{sec:game} of the current paper. In this section we only present our model of the literal listener and of the first-level pragmatic speaker.}} We show that, in this setting, vague messages can be, in some precise sense, optimal. This result relies on two substantive assumptions: first, that  the language's expressiveness obeys certain limitations, and, second, that the linguistic meaning of a vague expression contains an open parameter about which the listener is uncertain.}
%under the assumption that the language's expressiveness obeys certain limitations, semantically vague expressions can be more informative than their precise lexical alternatives. Limited expressiveness is a substantive assumption, but one that would not be explanatory of vagueness without the semantic definition that we combine it with
\pe{Conceptually, a key difference with Lipman's approach is that it rests on a semantic characterization of the notion of vagueness.}

\bs{This approach is limited in that it does not view the listener as an agent who reasons about the speaker's strategy, and therefore does not characterize the listener's and speaker's behavors as corresponding to an equilibrium of a signaling game. In fact, relative to the speaker's strategy, the literal listener's response to a message is not optimal. In section \ref{sec:game}, we provide a game-theoretic analysis and a method of equilibrium selection which will vindicate our core claim about the function of vaguenes and will allow us to prove a stronger result.}

% \pe{We show, under the assumption that the language's expressiveness obeys certain limitations, that semantically vague expressions can be more informative than their precise lexical alternatives. Limited expressiveness is a substantive assumption, but one that would not be explanatory of vagueness without the semantic definition that we combine it with.}

%Limited expressiveness is a substantive assumption, but one that would not be explanatory of vagueness without the semantic definition that we combine it with.}

%\pe{We deploy it with an assumption of limited expressiveness of the language in order to derive the fact that semantically vague expressions can be more informative than their precise lexical alternatives. This is a substantive assumption, but one that would not be explanatory of vagueness without the semantic definition that we combine it with.}
%it is in combination with the semantic characterization of vagueness that we give that it is explanatory.}

%does interact with our semantic characterization of vagueness in a way that is explanatory.}
%\pe{Another difference is that we limit the expressiveness of the language in certain ways.}
%The gist of the account is to explain that vagueness can be more informative than precision, by characterizing vagueness in terms of semantic openness.
%We first present the principles of the approach, and then illustrate it on an example.

\subsection{Vagueness as open meaning}\label{sec:openmeaning}

Let us start with a statement of the optimality result established in that paper. While not stated in that abstract form in \cite{egre2023optimality}, this result follows directly from the construction presented  there, which we explain further below. \pe{We call a \textit{context} a state of the world in which the speaker makes a private observation about the variable of interest.}\medskip

{\bf Proposition} ({\sc Optimality}): for some common prior distribution between a speaker $S$ and a listener $L$ on some variable of interest, {there exists a context $c$ and a vague sentence $m$}, % and a context $c$} 
such that relative to $S$'s information state, using $m$ provides more information to $L$ about the variable of interest than would any precise alternative $m'$ to $m$.\medskip

Several ingredients are involved in the proof of this fact. The first concerns the definition of vagueness. On this approach, vagueness is defined semantically, and not purely pragmatically, by the existence of an open semantic parameter affecting the truth conditions of the sentence in question. %By the same token, given a vague expression, the assumption is that it will include as lexical alternatives expressions that set this open parameter. 
Secondly, the result assumes a Bayesian model involving \bs{(a) a listener who is uncertain about some open semantic parameter and takes into account this uncertainty when interpreting vague messages, and (b) an uncertain speaker whose goal is to communicate their private probability distribution over some variable of interest.}
Thirdly, the result rests on a definition of comparative informativeness, defined using the notion of Kullback-Leibler divergence. Lastly, the account needs to specify a notion of precise alternative to a vague sentence. Let us consider these assumptions one by one.

\paragraph{Open meaning.} Consider a word like ``tall'' compared to ``taller than 185cm". Intuitively, the meaning of the first is open, it does not specify any height, whereas the second is closed, it does specify a height. In the literature on gradable adjectives, the difference is standardly cashed out by assuming the following truth conditions, in which $\mu_{height}(x)$ is a measure function returning the height of $x$ (see \citealt{bartsch1972grammar,kennedy2005scale,kennedy2007vagueness}):

\ex. \a. ``Gloria is tall'' is true iff $\mu_{height}(Gloria) \geq t$.
\b. ``Gloria is taller than 185cm'' is true iff $\mu_{height}(Gloria) \geq 185cm$.

On that approach, ``tall'' has an implicit comparative meaning, it means ``taller than $t$'', where $t$ denotes an open and context-sensitive threshold. Similarly, consider the word ``noonish'', compared to the word ``noon'', the truth conditions of the former can also be specified in terms of an open parameter: 

\ex. \a. ``The game starts at noon'' is true iff $time(game)=12pm$.
\b. ``The game starts noonish'' is true iff $time(game)=12pm\pm t$. 

The same contrast is in play with the preposition ``around'', compared to ``between'':

\ex. \a. ``Around 20 persons attended'' is true iff $x\in [20-t, 20+t]$.
\b. ``Between 10 and 30 persons attended'' is true iff $x \in [10,30]$.

Assuming $x$ is the number of people who attended, then the former sentence conveys that $x$ is in an indeterminate interval centered on 20, whereas the latter specifies a fixed interval. 

The assumption that vague expressions have open meanings coheres with a version of the supervaluationist conception of vagueness, which assumes that vague expressions express a multiplicity of more precise potential meanings (see \citealt{mehlberg1958reach,fine1975vagueness}).
{This characterization of vagueness in terms of openness in truth-conditions also makes vague expressions akin to indexical expressions (in the sense stated by \citealt[p.~68]{montague1970pragmatics}, of being a ``word or sentence one of which the reference cannot be determined without knowledge of the context of use''). But plenty of indexical expressions are not vague. A pronoun like ``I'' is indexical because its reference is context-sensitive. But its reference is fixed given the context: the rule that sets the reference as a function of context is rigid. For ``tall'', ``around'', ``noonish'', even setting the context, the meaning remains open. In that sense, ``I'' is a context-sensitive expression but it is not open in the way ``tall'' is.} 

\paragraph{Uncertainty.} On the speaker's side, our main pragmatic assumption to explain vagueness is that the use of a vague expression is going to be beneficial to speaker and to listener provided, first of all, the speaker herself is uncertain about the property in question (see \citealt{frazee2010vagueness}). For instance, if the speaker does not know Gloria's height with precision, or does not know if the game starts at exactly 12pm, or does not know the exact number of people who attended. If the speaker were not uncertain, then being cooperative she should communicate the exact quantity. 

When uncertain, using ``tall'', ``noonish'', or ``around'' will allow the speaker to minimize error in particular, in line with Grice's maxim of Quality (\citealt{grice1967logic}). But this is not the only function of vagueness, since the speaker could also choose to use an expression with precise but very weak truth-conditions, such as ``taller than 150cm'', or ``between 11am and 13pm", or ``between 0 and 50''. Our idea, rather, is that using a vague expression will also allow the speaker, in some cases, to give the listener more information about the shape of her uncertainty, an aspect that we link to Grice's maxim of Quantity.

On the listener's side, when a sentence containing a vague word is used as in our previous examples, {the listener has learned that a certain relationship holds between the variable of interest (e.g. someone's height) and the open parameter that is part of the meaning of the vague word. For instance, if I learn that Gloria is tall, assuming some relevant comparison class, I have learned that Gloria's height exceeds $s$, where $s$ is the threshold for tallness in that comparison class. Under uncertainty about both Gloria's height and the value $s$, I can use Bayes rule to obtain a posterior on both values, and from there, a new distribution on the possible values for Gloria's heights. The key observation is the following: if I don't know the value of $s$, then the higher $h$ is, the more likely it is that $h \geq s$. Therefore, by Bayesian reasoning, if I learn that Gloria is tall, i.e. that $h \geq s$, I should update my estimate of Gloria's height upward} (see below for details). 

When interpreting a vague statement, the task of the listener is therefore to figure out two things: on the one hand, Gloria's height/the time of the game/the number of attendees, and on the other hand, the value of the open semantic parameter appearing in ``tall''/``noonish''/``around 20''. Typically, given a vague predicate $\phi(x,t)$, where $t$ is the open semantic parameter, and $x$ is the variable of interest (height, time, number), {what the listener is really interested in is the value of $x$, which she tries to infer given the knowledge that whatever the `true' values of $x$ and $t$ are,  $\phi(x,t)$ is true}:

\ex. $P_L(x=k | \phi(x,t))$

This is where Bayesian inference comes in. We assume that the Listener $L$ has a joint prior distribution $P_L(x=k,t=i)$ on the variables $x$ and $t$ of interest. Our default assumption is that the two variables are probabilistically independent, so that: 

\ex. $P_L(x=k,t=i)=P_L(x)\times P_L(t)$.

For ``around'', for example, we assume independence between the listener's prior on how many attendees there might be, and the listener's prior on the intervals intended by ``around''. For ``tall'', it would be independence between the listener's prior on heights, and the listener's prior on the threshold intended by ``tall''. {Given such a prior joint distribution over the variable of interest and the open semantic parameter, the listener can compute the posterior distribution that results from updating with the information that $\phi(x,t)$ is true.}

\paragraph{Comparative informativeness.} In a cooperative language setting, we view the speaker's goal as being to bring the listener's posterior probability distribution on the variable $x$ of interest as close as possible to hers. Whether it concerns the use of precise or vague expressions, the goal of the Speaker is to be maximally informative, in line with Grice's maxim of Quantity, including about her own epistemic state. 

For that, we need an account of how well a probability distribution approximates another. The metric we use for that is given by the concept of Kullback-Leibler divergence (KL-divergence for short), also called relative entropy (see \citealt{kullback1951information} for the source, and \citealt{mcelreath2018statistical} for an introduction). Our model assumes a common prior distribution $P$ between speaker and listener on the variable $x$ of interest. We assume that the speaker first makes a private observation $P_o$, then sends a message $m$ to the listener. The listener then builds up a posterior distribution $P_m$ based on the message received. Using standard information-theoretic concepts, we can represent the speaker's surprisal regarding the value of $x$ after observing $o$ by the function $-\log(P_o(x=k))$, and the hearer's surprisal after getting the message $m$ by $-\log(P_m(x=k))$. The goal of the speaker is to minimize the average difference between the listener's surprisal and the speaker's surprisal after learning the true value of $x$. This is exactly what the KL-divergence formalizes, namely (assuming $x$ can only take discrete values for simplicity):
%$$D_{KL}(P_o||P_m) = \sum\limits_{k}P_o(x=k).\log (\frac{P_o(x=k)}{P_m(x=k)})$$

\begin{align*}
D_{KL}(P_o||P_m)=& \sum\limits_{k}P_o(x=k)[\log(P_o(x=k)) - \log(P_m(x=k))]\\
= &\sum\limits_{k}P_o(x=k)\log \left(\dfrac{P_o(x=k)}{P_m(x=k)} \right)
\end{align*}

Note that if $P_m$ were to assign 0 to a value that $P_o$ does not rule out, in case the speaker wrongly imparted that this value is excluded,  then $D_{KL}(P_o||P_m)$ would be \textit{infinite}. This would correspond to a \textit{violation of the maxim of Quality}, namely of the requirement of truthfulness from the speaker. In that regard, this metric combines the Quality and Quantity requirements of the Gricean approach. Finally, we assume that the speaker's utility of using a message $m$ after making an observation $o$ is given by $U_{S,m,o}=-D_{KL}(P_o||P_m)$

\paragraph{Lexical alternatives \pe{and expressiveness}.} We now have nearly all of the ingredients in place to establish that using vague sentences can be optimal over using precise sentences. We call a sentence vague if it contains a semantically vague word, and precise otherwise. What needs to be established is that in some cases, using a vague sentence will allow the speaker to achieve a KL-divergence that is strictly lower than the KL-divergence of any precise alternative sentence.
%lower the KL-divergence attached to a vague sentence compared to the KL-divergence attached to a precise alternative sentence. 
For the account to get off the ground, what is missing still is a specification of the notion of lexical alternative to a vague expression. 

The most natural assumption here is that the lexical alternatives are given by all the precise truth-conditional ways of answering a question under discussion. For ``tall'', for example, assuming the question at issue concerns Gloria's height, the precise alternatives would include all expressions that specify Gloria's height fully (``$n$ cm tall''), but also all the expressions that give a precise albeit partial indication of her height (``taller than $n$ cm'', ``shorter than $k$ cm'', and their conjunctions). For ``around'', this would similarly include all expressions specifying an exact value or a closed or semi-closed interval.

The specification of lexical alternatives is a key parameter here (as well as in other accounts of pragmatic selection of the lexicon, see the literature on scalar implicatures, \citealt{fox2011characterization,rothschild2013game}). In the case of ``tall'', for example, more alternatives could be included than just interval expressions, \pe{in particular} by incorporating explicit expressions of probability, such as ``$n$ cm tall with a probability of $\alpha$'' \pe{and their Boolean combinations}. As the reader might feel, however, these expressions are \pe{more complex and intuitively more cumbersome than an expression of the form ``around $n$''}. A full-fledged account ought to include such complex expressions, and to attach a notion of cost to them. Here, we follow \cite{egre2023optimality} and will assume that a sufficient argument for the optimality of vagueness can be established by restricting ourselves to lexical alternatives that do not incorporate explicit probabilistic talk. \pe{This restriction on expressiveness is a key assumption in our approach. Without it, the speaker could in principle state a full verbal description of their probability distribution, in a way that would count as maximally informative and as precise in the sense in which we defined that notion. Nevertheless, our point is that we can provide a context of limited expressiveness in which the use of a vague expression is optimal in a sense that is not trivial, and that is revealing of the function of vague words more broadly.} 

%We return to this issue below.

%in the discussion part.

%We return to this point later.

%For ``tall'', technically the alternatives are all the expressions of the form ``taller than $k$'', and ``of size equal to $k$''; for ``noonish'' these expressions include all the expressions denoting precise time intervals 

\subsection{Proof: the case of ``around''}

%Here is a scenario in which the use of the semantically vague approximator word ``around'', followed by a numeral $n$, would be optimal compared to more precise alternatives including 

To prove the optimality result stated above, it suffices to build a scenario in which the use of a semantically vague word is more informative than the use of any precise alternative. In \cite{egre2023optimality}, we consider the case of the approximator word ``around'' in numerical expressions of the form ``$x$ is around $n$'', whose truth conditions are open in the sense we explained. We take the precise lexical alternatives of a sentence like ``around $n$ persons attended'' to be all sentences of the form ``between $k$ and $m$ persons attended'', including cases in which $k=m$, in which case the expression is equivalent to ``exactly $k$ persons attended". The latter is in fact the most informative and obviously the most precise answer that could be given to the question at issue ``how many persons attended?".

\begin{figure}[t]
\[
\includegraphics[scale=.23]{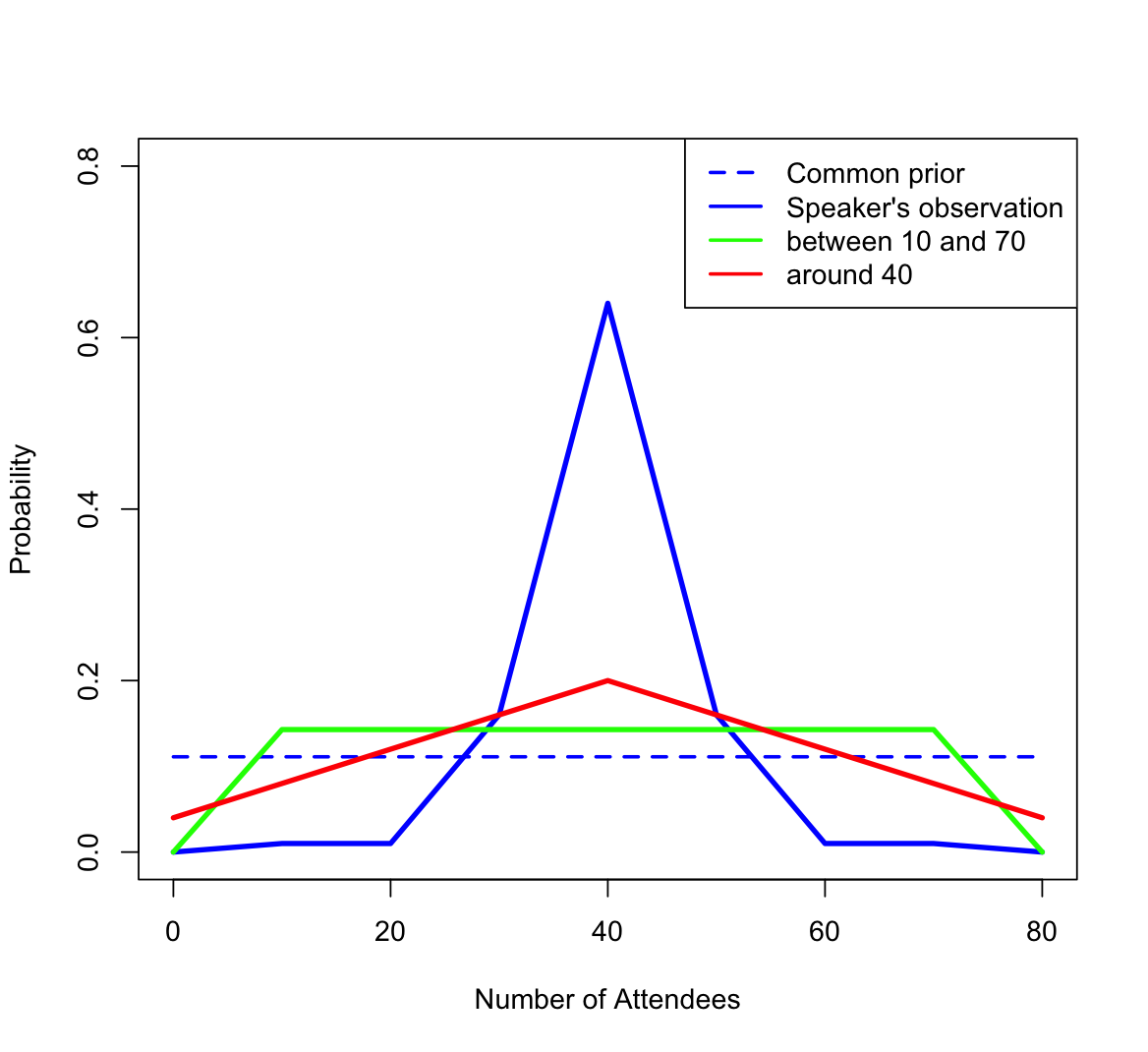}\]
\caption{``around'' vs ``between''}\label{fig:around}
\end{figure}

We assume that listener and speaker have a common uniform prior $P$ on the number of attendees, and to make the example very simple we also assume that attendees come in multiples of 10. The speaker made a private observation, and could rule out that either 0 or 80 persons attended, but is uncertain about other values, although her best guess would be 40, as reflected in the posterior $P_{o}$ (see row 2 of Table \ref{tab:around}, and the solid blue plot in Figure \ref{fig:around}).\footnote{\bs{A full implementation has to specify a set of possible observations and their own prior probabilities. Crucially, since the observation $o$ gives rise to the posterior distribution $P_o$ described in the second row of Table \ref{tab:around}, there must exists at least an other possible observation $o'$ giving rise to a very different posterior distribution, so as to ensure that the prior distribution on $k$ is indeed uniform. In particular, if there is only one other possible observation $o'$, we must have : for every $m$, $P(k=m) = P(o) \cdot P_o(k=m) + P(o') \cdot P_o'(k=m) = \nicefrac{1}{9}$. In Appendix \ref{sec:appendixexample}, we characterize a complete joint probability distribution over obsevations and the value of interest ($k$) such that one of the observations gives rise to a posterior distribution identical to the one shown in table \ref{tab:around} ($P_{o}(k)$), and such that the prior probability distribution over $k$, obtained by marginalizing over observations, is uniform on $[0,8]$.}}

\begin{table}[h] 
\caption{{\small Common Prior, Speaker's posterior, and Listener's posteriors for ``around'' vs. ``between''}}\medskip
\centering
\begin{tabular}{c|ccccccccc}
  \hline
$k$ & 0 & 10 & 20 & 30 & 40 & 50 & 60 & 70 & 80 \\ 

\hline\hline
$P(k)$ & \nicefrac{1}{9} & \nicefrac{1}{9} & \nicefrac{1}{9} & \nicefrac{1}{9} & \nicefrac{1}{9} & \nicefrac{1}{9} & \nicefrac{1}{9}& \nicefrac{1}{9} & \nicefrac{1}{9} \\
\hline\hline
  $P_{o}(k)$ & 0 & 0.01 & 0.01 & 0.16 & 0.64 & 0.16 & 0.01 & 0.01 & 0 
  \\ 
   \hline\hline
  $P_{\textrm{\emph{between}}}(k)$ & 0 & \nicefrac{1}{7} & \nicefrac{1}{7} & \nicefrac{1}{7} & \nicefrac{1}{7} &  \nicefrac{1}{7} &  \nicefrac{1}{7} &  \nicefrac{1}{7} & 0 \\
  $P_{\textrm{\emph{around}}}(k)$ & 0.04 & 0.08 & 0.12 & 0.16 & 0.20 & 0.16 & 0.12 & 0.08 & 0.04 \\ 
\end{tabular}\label{tab:around}
\end{table}

By hypothesis, the speaker can choose a variety of messages, using either ``between'' or ``around''. For a Bayesian update with ``between $k$ and $m$'', the listener's posterior is computed simply by spreading the mass initially allocated to numbers outside the interval $[k,m]$ to that interval. Starting from a uniform prior on the number of attendees, this produces a uniform posterior, $P_L(x=k \mid \text{$x$ is between $m$ and $m'$}) = \frac{1}{|m'-m+1|}$. For a Bayesian update with ``around $k$'', assuming the prior is uniform both on the number of attendees and on the meaning of ``around'', that the support of $x$ is the interval $[0,2n]$ and that of $t$ in ``around'' is $[0,n]$, then under the assumption of independence stated above, it can proved that: $P_L(x=k \mid \text{$x$ is around $n$}) = \frac{n-|n-k|+1}{(n+1)^2}$ (see \citealt{egre2023optimality} for details, and for the general case of arbitrary priors).

%\begin{table}[h]
%\caption{Listener's posteriors after hearing  ``between 1 and 7'' and ``around 4''.\label{Pm}}\medskip
%\centering
%\begin{tabular}{r|ccccccccc}
%  \hline
%$k$ & 0 & 10 & 20 & 30 & 40 & 50 & 60 & 70 & 80 \\ 
%\hline
% $P_\textrm{\emph{between}}(k)$ & 0 & \nicefrac{1}{7} & \nicefrac{1}{7} & \nicefrac{1}{7} & \nicefrac{1}{7} &  \nicefrac{1}{7} &  \nicefrac{1}{7} &  \nicefrac{1}{7} & 0 \\
%  $P_\textrm{\emph{around}}(k)$ & 0.04 & 0.08 & 0.12 & 0.16 & 0.20 & 0.16 & 0.12 & 0.08 & 0.04 \\ 
%   \hline
%\end{tabular}\label{tab:post}
%\end{table}

Using both formulas, we obtain the listener's posteriors produced in Table \ref{tab:around} for ``between 10 and 70'' and for ``around 40'', which we notate $P_{around}$ and $P_{between}$. Neither posterior is identical to the speaker's distribution $P_{o}$, but $P_{around}$ yields a lower KL-divergence, since $D_{KL}(P_{o}||P_{\textrm{\emph{between}}}) = 0.89$, while $D_{KL}(P_{o}||P_{L,\textrm{\emph{around}}}) = 0.65$.  This suffices to establish our result, since the most informative ``between''-message the speaker can use is ``between 1 and 7'', as other choices of bounds would lead the listener to rule out values that the speaker thinks might be actual, yielding an infinite KL-divergence. Additionally, it is also the case that other choices of the target value than $n=40$ for ``around $n$'' would be less good in this case (such as ``around 30'', ``around 50'', etc), for they would yield a higher KL divergence too.

In summary, in this scenario, the speaker's utility is strictly greater for the ``around 40'' message than for any ``between'' message, which implies that the speaker should strictly prefer the ``around'' message. Although the latter does not coincide with the speaker's observational distribution, it is peaked like it, tailed like it, and it provides a better approximation than any truth-conditionally precise counterpart (see Figure \ref{fig:around}).

\subsection{Generalizing to other vague words
}\label{sec:generalizing} % : an example with ``tall''

%Further work needs to be done to explain vagueness. \BS{however, and several difficulties remain.}{} \BS{First of all,}{In particular,} 

The optimality result we stated above makes a weak claim. It shows that at least one semantically vague expression can have optimal use over its precise lexical alternatives, but not that all of them do. However, we reckon that the same strategy we used for ``around'' could be used to establish the following stronger claim involving universal quantification:\bigskip

\textbf{Hypothesis ({\sc General optimality}):} For every semantically vague term $V$, speaker $S$ and listener $L$, there is a common prior distribution on some variable of interest and a context $c$, such that relative to $S$'s information state, using the term $V$ is more informative to $L$ about the variable of interest in $c$ than any semantically precise alternative $V'$ to $V$ {(barring explicit probabilistic talk).}\bigskip

\noindent The reason this appears to be a reasonable claim is that the statement of this hypothesis involves several degrees of freedom, both regarding the choice of the common prior, and then regarding the speaker's information state and its associated probability distribution.

%shape of the speaker's distribution. 

\begin{figure}[t]
\[
\includegraphics[scale=0.22]{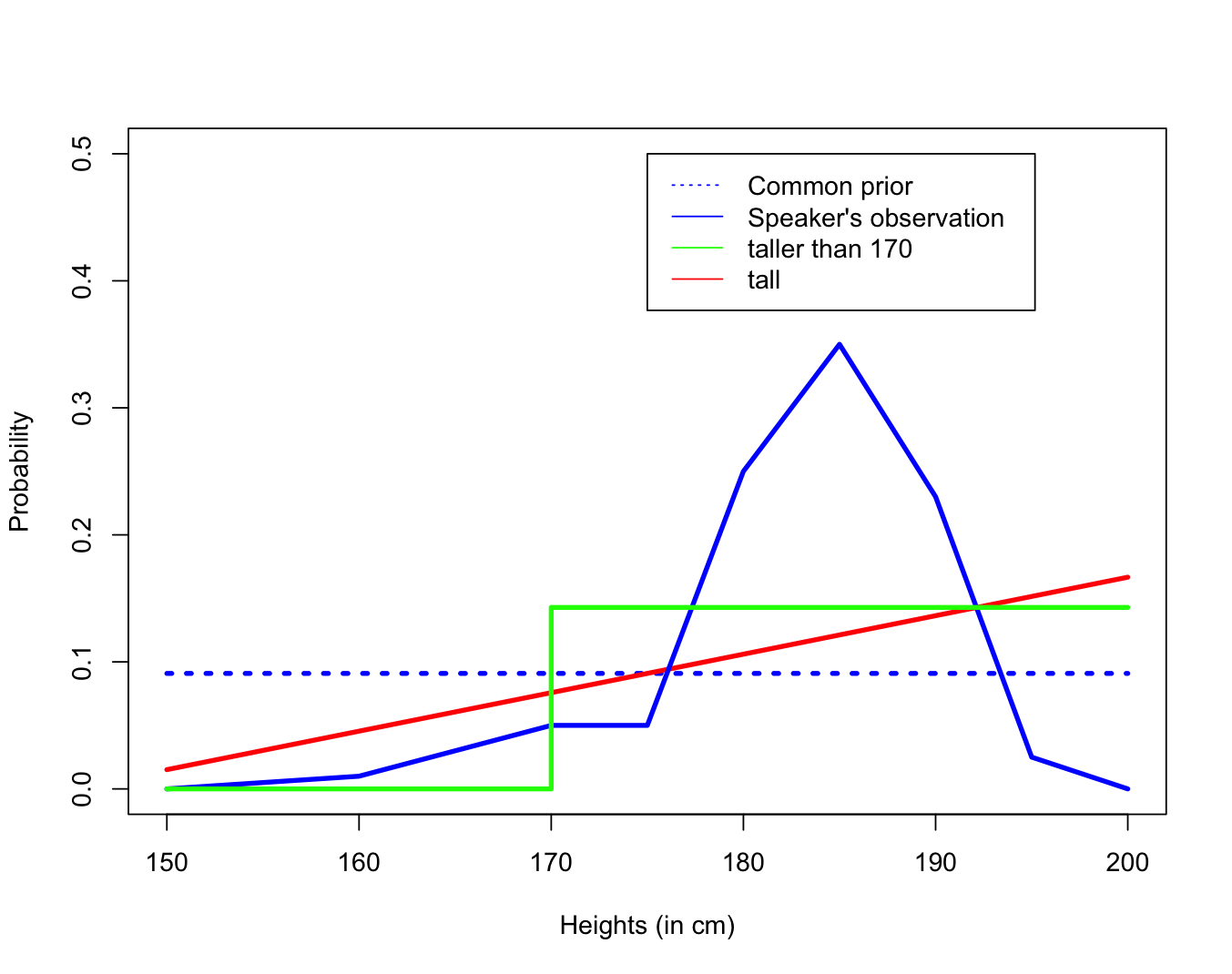}
\]
\caption{``tall'' vs ``taller than''}\label{fig:tall}
\end{figure}

To see how such a generalization might go, consider the case of ``tall'', with the assumption that it means ``taller than $t$''. Let us give ourselves the same assumptions we used for ``around'', namely that speaker and hearer start from a common uniform prior on possible heights of a variable of interest, that the listener also has a uniform prior regarding the threshold values that ``tall'' can take, and that both speaker and hearer restrict the domain to the range 150-200cm.\footnote{%The assumption of a uniform prior on heights is obviously irrealistic and we could assume a Gaussian prior instead. 
We keep to uniform priors first to make the case maximally simple and comparable to our treatment of ``around''. See the next page, and \cite{lassiter2017adjectival}, for examples of ``tall'' involving Gaussian priors on heights.} For simplicity, we again assume that heights vary discretely by steps of 5cm. Under those assumptions, the listener's posterior after hearing ``tall'' can be computed using Bayes' Theorem, and it has the property that higher heights become comparatively more probable than smaller heights (see the red plot in Figure \ref{fig:tall}).\footnote{When the range of $x$ is $[0,n]$ and likewise for the threshold parameter $t$ in ``tall'', then it can be proved analytically that $Pr_L(x=k|x\geq t)=\frac{2(k+1)}{(n+1)(n+2)}$, which is the linear equation of the red plot, mapping the interval $[150,200,by=5]$ to $[0,10,by=1]$).}

If the question at issue is ``what is Gloria's height?'', then ``tall'' admits various precise lexical alternatives, all alternatives selecting bounded height intervals, such as ``between $k$ and $j$ cm'' or ``more than/less than $k$ cm''. %% calculate the posterior based on the uniformity assumption
Figure \ref{fig:tall} presents a case in which the speaker's private observation follows a distribution peaked on 185cm, where the very extreme values have probability 0.
%, with the property that the extreme values 150 and 200 do not receive probability 0 but a probability very close to zero. 
%In this case, all expressions denoting intervals (``between $n$ and $m$'',``taller than $m$'',``shorter than $m$'') yield a posterior that gives a zero value for some value in the speaker's observational support. Hence, the KL-divergence to the speaker's observation is infinite, which rules out such messages (one of those distributions is represented by the step function in green for ``taller than 170cm'' in Figure \ref{fig:tall}). 
In this case, the posterior for ``tall'' produces a smaller KL-divergence than the posterior produced by expressions denoting precise intervals. Such expressions either give a zero value in the speaker's observational support, and therefore produce an infinite KL-divergence (such an interval is shown by the step function in green for ``taller than 170cm'' in Figure \ref{fig:tall}), or they give a finite KL divergence (``between 155 and 195cm'', ``more than 155cm'', ``less than 195cm'') that is still higher than the divergence produced by ``tall'' in this toy example. However imperfect the linear update for ``tall'' turns out compared to the speaker's observational distribution, it comes closer to it in this case. 

%The effect is obviously more general: if the speaker's observation is sufficiently biased in favor of higher values, using \textit{tall} might be more informative than using an interval of the form ``between'' whose support corresponds to that of the posterior. Figure \ref{fig:tall_normal} shows the case of posterior for ``tall'' generated from a Gaussian prior on heights this time.%Such a case, involving a Gaussian prior on heights this time, is represented in Figure \ref{fig:tall_normal}.

%relative to the speaker's observational distribution than the prior, corresponding to the trivial message ``between 150cm and 200cm''. So again, we have found a context and a distribution that rationalizes the use of a vague word like ``tall'', relative to a set of precise alternatives. 

\begin{figure}[t]
\[
\includegraphics[scale=0.22]{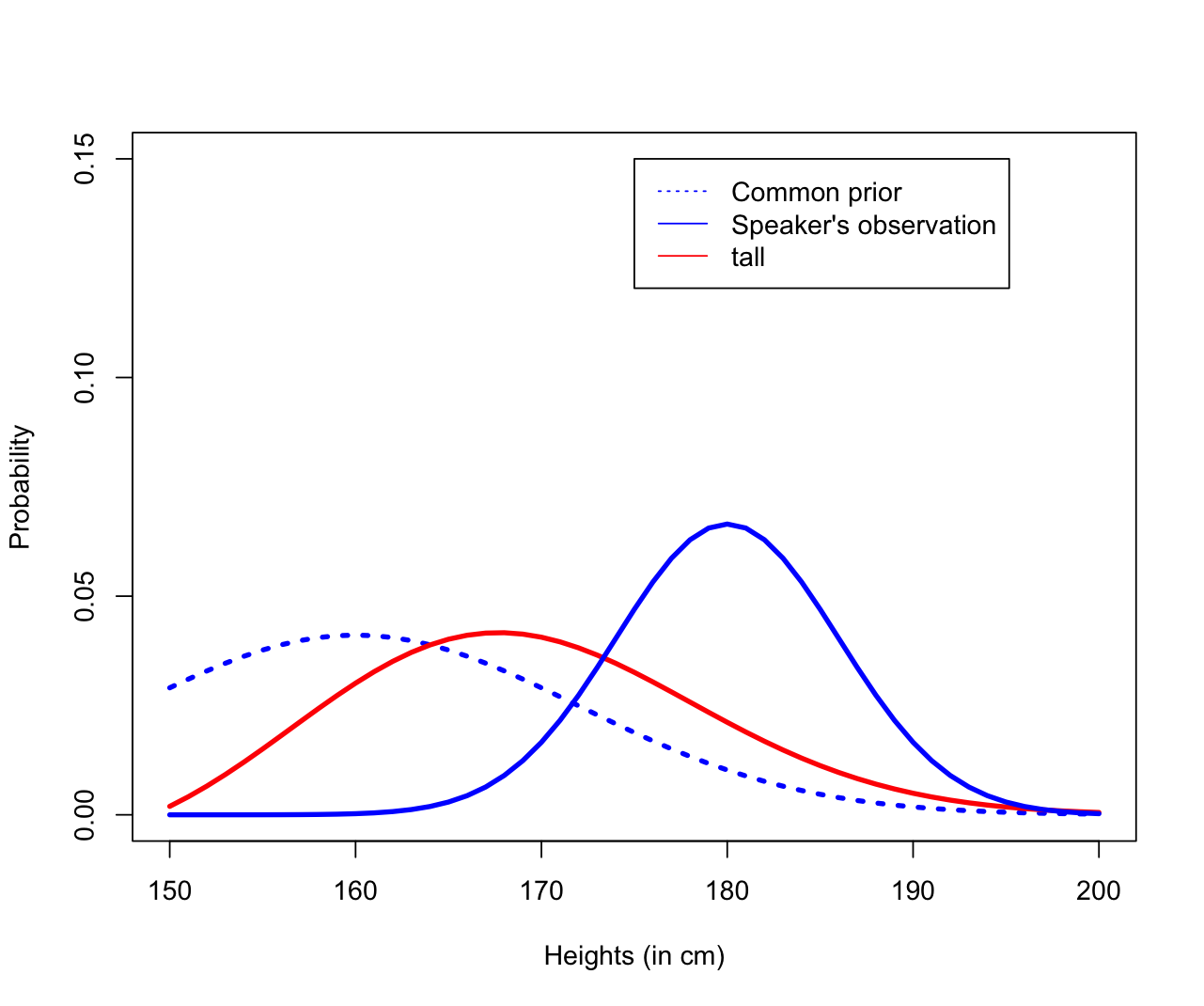}
\]
\caption{Posterior on heights resulting from ``tall'', starting from a Gaussian prior}\label{fig:tall_normal}
\end{figure}

%Of course, there will also be contexts in which, if the speaker can categorically rule out extreme values in the range 150-200cm, the use of a precise ``between'' message might have a lower KL-divergence than the use of the ``tall'' message. {Even in such a case, however, if the speaker is extremely biased in favor of higher values, using \textit{tall} might be more informative than using ``between 150 and 200''.} Such a case, involving a Gaussian prior on heights this time, is represented in Figure \ref{fig:tall_normal}.

Quite generally, it can be proved that using ``tall'', given its truth-conditional content, will shift the listener's prior on heights toward higher values, irrespective of the shape of the listener's prior. More precisely, irrespective of the prior on heights and of the prior on thresholds, we have, when $k_2 > k_1$, the following Ratio Inequality:
%\footnote{
%Proof.
%\begin{align*}
%Pr(x=k_2|x\geq t) &= \frac{\sum_{k_2\geq i} Pr(x=k_2)\times Pr(t=i)}{Pr(x\geq t)}, \text{and}\\
%Pr(x=k_1|x\geq t) &= \frac{\sum_{k_1\geq i} Pr(x=k_1)\times Pr(t=i)}{Pr(x\geq t)},\text{hence} \\ 
%\dfrac{Pr(x=k_2|x\geq t)}{Pr(x=k_1|x\geq t)}&=\frac{\sum_{k_2\geq i} Pr(x=k_2)\times Pr(t=i)}{\sum_{k_1\geq i} Pr(x=k_1)\times Pr(t=i)}\\
%&=\frac{\sum_{k_2\geq i} Pr(t=i)}{\sum_{k_1\geq i} Pr(t=i)}\times \dfrac{Pr(x=k_2)}{Pr(x=k_1)}.\\
%\text{However, }
%\frac{\sum_{k_2\geq i} Pr(t=i)}{\sum_{k_1\geq i} Pr(t=i)}&=\frac{\sum_{k_1\geq i} Pr(t=i)+\sum_{k_1<i\leq k_2}Pr(t=i)}{\sum_{k_1\geq i} Pr(t=i)}> 1.\\
%\end{align*}
%}
%This fact follows from the same ratio inequality mentioned in the case of ``around''. 

$$\dfrac{Pr(x=k_2|\text{$x$ is tall})}{Pr(x=k_1|\text{$x$ is tall})} > \dfrac{Pr(x=k_2)}{Pr(x=k_1)}$$

\noindent The proof of this Ratio Inequality is simple and given in the appendix. It is similar to the proof given in \cite{egre2023optimality} to show that after hearing ``around $n$'', values closer to $n$ receive comparatively more weight than values further away from $n$ (as visible on Figure \ref{fig:around}). Figure \ref{fig:tall_normal} shows the case of a posterior for ``tall'' generated from a Gaussian prior on heights, illustrating the same effect, whose non-linear shape this time approximates the speaker's observation much closer.

%% cite Lassiter and Goodman; use the ratio inequality to prove the fact

Given that fact, we can therefore understand the the rationality of using ``tall'' as the fact that precise interval expressions will in many cases produce posteriors whose shape is too rigid to approximate the speaker's observational distribution in a satisfactory way. Vague words, seen through that lens, thus appear as expressive resources allowing us, in some contexts, to more flexibly communicate the shape of our inner representations of the world than their more rigid precise counterparts.

%\color{violet}

\section{A game-theoretic analysis, and a global optimality result}\label{sec:game}

A major limitation of the model we have presented is that the listener is not strategically rational with respect to the speaker: the listener's interpretation of a message -- that is, her posterior beliefs -- is based solely on the literal meaning of the message, without taking into account how the speaker chooses messages. A listener who is aware of the speaker's strategy should condition her interpretation on this knowledge. In other words, the pair consisting of the listener's interpretation function and the speaker's production strategy does not constitute a signaling-game equilibrium.

In this section, we provide a game-theoretic model in which this limitation is overcome. Inspired by Franke's Iterated Best Response model, we define a specific equilibrium concept (the IBR equilibrium) where the speaker's and listener's strategies are anchored in the literal meaning of messages (see \citealt{franke2009signal}).\footnote{\bs{Our model is different from Franke's in that Franke's utility function for speakers is not based on Kullback-Leibler Divergence, and Franke's listeners have their own utility function based on actions that they take in different situations. From this perspective, our approach is closer to the Rational Speech Act approach of \cite{goodman2013knowledge} (RSA for short),  except that in the RSA approach, the speaker is not modeled as fully rational, as a result of which the predicted behavior of speakers and listeners does not constitute a  Nash equilibrium. Franke's model is itself related to previous game-theroretic work that is concerned with best response dynamics and iterated best response reasoning, such as  \cite{gilboa1991social,monderer1996potential,crawford2007level}, among many others. \cite{blume2025meaning} too makes use of an Iterated Best Response  mechanism in order to select equilibria which are anchored in the literal meaning of messages.}} 

  % (see also RSA + Blume \REF).
 We go on to prove a more global optimality result: in our setup, in the absence of vague messages, the speaker cannot communicate any information about their probability distribution over the variable of interest beyond its support.

\subsection{General set-up}

We consider the following signaling game. \pe{As we specify it here, this game assigns actions and payoffs only to the speaker. The listener's behavior is fully specified by the assumption that the listener is a Bayesian updater.} 

\subsubsection{Players, messages, strategies and payoffs}

\begin{itemize}
  \item \textbf{Players.}  A \emph{speaker} $S$ and a \emph{listener} $L$.

  \item \textbf{Messages.}  $S$ can send one of the finitely many sentences
        $M=\{m_1,\dots,m_n\}$.\footnote{\bs{Languages are infinite, but formal
        models often restrict attention to a finite message set for simplicity.}}

  \item \textbf{Worlds and observations.}  
        Nature chooses a world $w$ and an observation $o$ according to a known
        joint prior $P(w,o)$.  (We assume finitely many observations; the
        finiteness of worlds is not needed below.)

  \item \textbf{Timing.}
        \begin{enumerate}
          \item Nature draws $(w,o)$ according to $P$.
          \item $S$ observes $o$ and sends a message $m\in M$.
          \item $L$ observes $m$ and \emph{only} updates her belief about
                $(w,o)$; she takes no further action.
        \end{enumerate}

  \item \textbf{Speaker strategy.}  
        $S(m\mid o)$ is the probability of sending $m$ when $o$ is observed;
        thus a strategy is a map $S:O\to\Delta(M)$ from observations to probability distributions on messages.

  \item \textbf{Listener belief.}  
        If $m$ is received, let $L_m$ be the listener's posterior distribution on
        $(w,o)$.

  \item \textbf{Speaker payoff.}
        After $(w,o,m)$ is realised, the speaker's utility is
        \[
           U(m,o) \;=\; -\,D_{\mathrm{KL}} \bigl(P_o \,\|\, L_m\bigr),
           \qquad
           P_o(\cdot)=P(\cdot\mid o).
        \]
        In words, the speaker wants to minimize the KL-divergence  of her own distribution over $(w,o)$ from the listener's posterior joint distribution, hence to maximize the utility just defined, obtained by adding a minus sign.\footnote{\bs{Note that we consider here the joint distributions over $(w,o)$, rather than the marginal distributions over $w$. On the speaker's side, of course, $P_{o_i}(w,o|o_i) = 0$ whenever $o \neq o_i$. In \cite{egre2023optimality}, we used instead the marginal distributions over $w$. Here, we model the speaker as wanting to convey the observation she made, on top of wanting the listener to have a distribution over $w$ that is maximally close to her's. \label{joint}}}
             \end{itemize}

\subsubsection{Best response and equilibrium}

Let $S$ be a speaker strategy and $L$ a listener profile
($m\mapsto$ distribution on $(w,o)$).

\paragraph{Speaker best response.}
$S$ is a \emph{best response} to $L$ iff, for every observation~$o$,
\[
   S(m\mid o) > 0
   \;\Longrightarrow\;
   m\in\arg\!\max_{m'\in M}
         \bigl\{-\,D_{\mathrm{KL}}(P_o \,\|\, L_{m'})\bigr\}.
\]

\paragraph{Bayesian listener.}
If the listener believes the speaker uses $S$, then for any message $m$
with positive ex-ante probability
$S(m)=\sum_o P(o)\,S(m\mid o)>0$, she must set
\[
   L_m(w,o)\;\propto\; P(w,o)\,S(m\mid o)
   \quad\text{(Bayes’ rule).}
\]
(When $S(m)=0$, $L_m$ may be arbitrary.)

\paragraph{Equilibrium.}
A pair $(S,L)$ is an \emph{equilibrium} if
\begin{enumerate}
  \item $S$ is a best response to $L$ (condition above), and
  \item $L$ is Bayesian relative to $S$ (Bayes’ rule whenever $S(m)>0$).
\end{enumerate}
That is, the speaker never deviates from KL--optimal messages, and the
listener's beliefs are consistent with the speaker's actual strategy.

\subsubsection{Technical note: messages with $-\infty$ payoff and uninformative equilibria}

\noindent
Whenever $L_m$ assigns zero probability to a world $w$ that lies in the support of $P_o$ (i.e., is compatible with $o$),
the KL divergence is $+\infty$ and the speaker's utility is $-\infty$.
This is intentional: such messages are ruled out in equilibrium.

\medskip
\noindent
A potential issue is that, for some observation~$o$, the speaker might assign infinite KL divergence to \emph{every} message --- that is, $D_{\mathrm{KL}}(P_o \| L_m) = \infty$ for all $m \in M$. In such a case, the speaker would be indifferent across messages, and the best-response condition would become vacuous.

In fact, this situation can never arise at equilibrium under the current assumptions. Since the speaker's strategy must satisfy $\sum_m S(m \mid o) = 1$, there exists at least one message $m^*$ such that $S(m^* \mid o) > 0$. By Bayes’ rule, the listener's posterior for that message is given by
\[
L_{m^*}(w,o) = \frac{P(w,o)\,S(m^* \mid o)}{\sum_{w',o'} P(w',o')\,S(m^* \mid o')}.
\]
Whenever $P(w,o) > 0$, the numerator and denominator are both strictly positive, so $L_{m^*}(w,o) > 0$. It follows that $L_{m^*}$ assigns positive probability to every world the speaker considers possible under $o$, and therefore $D_{\mathrm{KL}}(P_o \| L_{m^*}) < \infty$. Thus, the speaker always has at least one message with finite utility.

\medskip
\noindent
This, however, does not rule out completely uninformative equilibria. A listener who interprets every message by returning the prior distribution, coupled with a speaker who randomizes identically across all messages for every observation, constitutes an equilibrium, but one in which no information is ever communicated.

\subsection{IBR equilibria: anchoring equilibria in linguistic meaning}

In general, there are many equilibria. Following \cite{franke2009signal}, we introduce one way to select a specific equilibrium, rooted in the literal meaning of messages. The idea is to start with a literal listener $L_0$ who interprets messages literally, then to consider a speaker $S_1$ who responds optimally to $L_0$. Specifically, $S_1$'s probability of choosing a message $m$ is $0$ if $m$ is not optimal, and is otherwise uniform over optimal messages (other choices could be possible regarding the choice of optimal messages). We can then define a more sophisticated listener $L_1$ who updates her belief based on Bayes' rule after receiving a message on the assumption that the speaker was  $L_1$, and then a speaker $S_2$ whose probability of choosing a message is uniform over all optimal messages relative to $L_1$, and so on and so forth. 

At each level $n$, we make a special provision for `surprise messages', i.e. messages which have a null probability of being used by $S_n$: such messages give rise to the same interpretation as when they are interpreted by $L_0$ (\citealt{franke2009signal}). If the sequence $(L_n,S_n)$ converges, its limit $(L_{+\infty}, S_{+\infty})$ will satisfy the conditions above, as will be shown below, hence will be an equilibrium. We call this equilibrium, if it exists, the IBR equilibrium. \\
\\
 The literal meaning of a message $m$ is a function that takes two arguments: a world $w$ and \pe{an assignment $g$ of values to the open parameters of $m$} (viz. threshold values for all gradable adjectives). %values for 
 %a set of open parameters $i$ ($i$ includes for instance threshold values for all gradable adjectives). 
 We write $\sem{m}^{w,g} = 1$ when, %given a choice $i$ for open parameters, 
 $m$ is true in $w$; we write $\sem{m}^{w,g} = 0$ otherwise. The limited expressiveness assumption discussed in section \ref{sec:account} is enforced here, from the mere fact that the truth-value of a message depends on $w$ and $g$, but not on $o$. In other words, the language is assumed to contain propositions that are fully `objective', i.e. contain information about the variable of interest, but no information about the private observation made by the speaker. 
 
 $P$ is a joint distribution over $o$, $w$, and $g$, where $g$ is independent of $w$ and $o$. Finally, we assume that among the messages, there is a special \emph{null} message $m_\top$ which always returns $1$ (for every $w$ and every $g$). This ensures that, at every level of recursion, for every observation, there is at least one message with finite utility. 
Importantly, we define $L_0$ as taking into account the uncertainty about open parameters when interpreting messages. \\
 \\
 We now recursively define the sequence $(L_n, S_n)$. 
 \\
 \begin{enumerate}
\item 
\[
L_0(w,o \mid m) = \sum_g P(w,o,g \mid \sem{m}^{w,g} = 1)  
\propto P(w,o) \times \sum_{\{g : \sem{m}^{w,g} = 1\}} P(g).
\]

\item 
\[
U_k(m,o) = -D_{\mathrm{KL}}\big(P(\cdot \mid o) \,\|\, L_{k-1}(\cdot \mid m)\big),
\]
where \( P(\cdot \mid o) \) and \( L_{k-1}(\cdot \mid m) \) are the posterior joint distributions over worlds and observations, resulting respectively from observing \( o \) (for the speaker) and from processing message \( m \) (for the level-\((k-1)\) listener).

\item 
\[
S_k(m|o) =
\begin{cases} 
    0 & \text{if } m \not\in \operatorname*{argmax}_{m'} U_k(m',o), \\
    \dfrac{1}{|\{m' : m' \in \operatorname*{argmax}_{m'} U_k(m',o)\}|} & \text{otherwise}.
\end{cases}
\]

\item 
For \( k > 0 \),
\[
L_k(w,o \mid m) =
\begin{cases} 
    L_0(w,o \mid m) & \text{if } S_k(m) = 0\\
    & \text{(where } S_k(m) = \sum\limits_{o} P(o) \cdot S_k(m \mid o)\text{)}, \\
    \dfrac{P(w,o) \cdot S_k(m \mid o)}{\sum\limits_{w',o'} P(w',o') \cdot S_k(m \mid o')} & \text{otherwise}.
\end{cases}
\]

 \end{enumerate}
 
 \noindent We can now state the following result:

\vspace{1em}

\textbf{Proposition.} \textit{The IBR limit, if it exists, is an equilibrium.}

\vspace{1em}

\textbf{Proof.}  
Let \( f \) be the function that takes a listener profile \( L \) and returns a speaker strategy \( S \) such that, for every message \( m \) and every observation \( o \),
\[
S(m \mid o) = 
\begin{cases}
0 & \text{if } m \not\in \operatorname*{argmax}_{m'} U_k(m', o), \\
\frac{1}{|\{ m' : m' \in \operatorname*{argmax}_{m'} U_k(m', o) \}|} & \text{otherwise}.
\end{cases}
\]
Let \( f' \) be the function that maps a speaker strategy \( S \) to a listener profile based on Bayes' rule, together with the special rule for \emph{surprise messages} mentioned above (i.e., interpreting them according to their literal meaning). Note that for any listener profile $L$, $f(L)$ is a best response to $L$, and that for every speaker strategy $S$, $f((S)$ is Bayesian relative to $S$. Then, for every \( k \), we have:
\[
S_{k+1} = (f \circ f')(S_k).
\]
The function \( f \circ f' \) is not, in general, continuous (because  $\operatorname*{argmax}$ is not continuous, and because of the treatment of surprise messages), and so it is not immediately clear that the IBR limit must be a fixed point of this function.

To show that the limit \( (S_{+\infty}, L_{+\infty}) \), if it exists, is an equilibrium, we first show that this limit is in fact reached in a finite number of steps, rather than merely approached asymptotically. That is, there exists some \( r \in \mathbb{N} \) such that, for all \( r' \geq r \), we have \( S_{r'} = S_{+\infty} \) and \( L_{r'} = L_{+\infty} \).

Assume that the sequence \( (S_k)_{k \in \mathbb{N}^+} \) converges. Let \( N \) be the number of messages. For any message \( m \), observation \( o \), and rank \( r \), the value \( S_r(m \mid o) \) belongs to the finite set \( \{0, \frac{1}{N}, \frac{1}{N-1}, \ldots, 1\} \). Therefore, for any \( r \), either
\[
S_{r+1}(m \mid o) - S_r(m \mid o) = 0,
\quad \text{or} \quad 
S_{r+1}(m \mid o) - S_r(m \mid o) \geq \left( \frac{1}{N-1} - \frac{1}{N} \right).
\]
Since \( (S_k(m \mid o))_{k \in \mathbb{N}^+} \) converges, the sequence of differences \( (S_{k+1}(m \mid o) - S_k(m \mid o))_{k \in \mathbb{N}^+} \) must converge to 0. Because any non-zero element of this sequence is greater than \( \frac{1}{N-1} - \frac{1}{N} \), the sequence can only converge to 0 if it is equal to 0 after a certain rank.

That is, there exists some \( r \in \mathbb{N} \) such that for all \( r' \geq r \), we have \( S_{r'+1}(m \mid o) = S_{r'}(m \mid o) = S_{+\infty}(m \mid o) \). Therefore, for every message \( m \) and observation \( o \), the sequence \( (S_k(m \mid o))_{k \in \mathbb{N^+}} \) is constant after a certain rank. Since there are finitely many messages and observations, the entire strategy \( (S_k)_{k \in \mathbb{N^+}} \) is constant after a certain rank. Moreover, since \( L_k \) is fully determined by \( S_k \), the sequence \( (L_k)_{k \in \mathbb{N^+}} \) is also constant after a certain rank.

Thus, there exists a rank \( r \) such that for all \( r' \geq r \), we have, for all $m$, $w$ and $o$:
$$ S_{r'}(m \mid o) = S_{r}(m \mid o) = S_{+\infty}(m \mid o)$$
$$\text{and}\ L_{r'}(w, o \mid m) = L_r(w, o \mid m) =  L_{+\infty}(w,o \mid m)$$

Since \( S_{r+1} = S_r \) and since \( S_{r+1}  = f(L_r) \), $S_r$ is a best response to \( L_r \).  Moreover, \( L_r \) is a Bayesian listener strategy with respect to \( S_r \), since $L_r=f'(S)$. It follows that \( (S_r, L_r) \), which is the same as \( (S_{+\infty}, L_{+\infty}) \),  is an equilibrium. Hence the IBR limit, if it exists, is an equilibrium.

\smallskip \hfill ${\small \sf QED}$\bigskip

%\vspace{1em}

%\textit{Note: 

Note that the proof relies, among other things, on the assumption that the number of messages and the number of observations is finite.

\subsection{Global Optimality Result}

%\noindent \textbf{Definition} ({\sc Vagueness}) 

We say that a message $m$ is \emph{vague} provided it leaves open some interpretative parameters. Hence, a message is vague if there exists at least one world $w$ and \pe{two assignments $g_1$ and $g_2$ of values to the open parameters of $m$} such that %open parameter sets  $i_1$ and $i_2$ such 
that $\sem{m}^{w,g_1} \neq \sem{m}^{w,g_2}$. Remember that two observations $o_1$ and $o_2$ have the same support if for ever $w$, $P(w|o_1) > 0 \Leftrightarrow P(w|o_2) >0$. It can then be proved that:
\paragraph{Theorem.} In a signaling game within the set-up described above, if the IRB equilibrium exists and there is no vague message, then for any two observations $o_1$ and $o_2$ which have the same support,  at equilibrium, $S(\cdot \mid o_1) = S(\cdot \mid o_2)$ (i.e., for any message $m$, $S(m \mid o_1) = S(m \mid o_2)$).\\ 
\\
\pe{We give the proof of the Theorem in Appendix \ref{sec:appendixproof}.}\footnote{\bs{This result crucially relies on the specific utility function we use. In particular, if instead of defining the speaker's utility function in terms of the KL divergence of the speaker's \emph{joint} distribution over $(w,o)$ from that of the listener's, we only used the marginal distributions over $w$, as we  did in \cite{egre2023optimality}, the theorem would not hold. That is, it is crucial that we model the speaker as wanting to also convey the observation she made, and not just to make the listener's distribution over $w$ maximally close to her's. Cf. footnote \ref{joint}.}}
 Conversely, as shown in Appendix \ref{sec:appendixexample}, one can exhibit an example of a signaling game within our setup and which includes a vague message such that there exist two observations $o_1$ and $o_2$ such that, at equilibrium, $S(\cdot \mid o_1) \neq S(\cdot \mid o_2)$.\\
\\
Within this setup, therefore, the existence of vague messages is necessary for speakers to be able to communicate about the \emph{shape} of their distribution, not just their \emph{support}. This result is much stronger that the one produced in Section \ref{sec:openmeaning}. It does not just show that, in a given language, vague messages are sometimes optimal. Rather it shows that, under certain assumptions, the existence of vague messages is necessary in order to communicate fine-grained information about one's probability distribution. Of course, this result crucially relies on a strong limitation on the language's expressiveness, and more generally on the specific assumptions built in our model (the specific utility function used, for instance, the IRB dynamics, etc.). Despite these limitations, this result at least provides a plausible %, if speculative 
answer to the question why vagueness exists in natural languages. 

\color{black}

\section{Comparison: Semantic and Game-Theoretic Meaning}\label{sec:discussion}

%Now that we have a reasonably clear view of each account of vagueness, we can explain the sense in which both accounts are compatible. Imagine that the speaker picks the unique message $m$ that minimizes $D_{KL}(P_o||L_m)$ (if there is one), and in case of ties, invariably picks the around-message.

%This speaker's strategy is a pure strategy: relative to each observation $o$, this speaker chooses a certain message with probability 1. So there would be no vagueness according to Lipman's notion of vagueness.

Now that we have a clear understanding of both Lipman's suboptimality result and our own optimality results, we can explain the sense in which both accounts are compatible, and then address some broader meta-semantic claims made by Lipman in his paper.

Let us first establish the compatibility between both accounts. Consider the example involving the use of ``around''. Suppose that for each observation $o$, the speaker picks the unique message $m$ that minimizes $D_{KL}(P_o||P_m)$ (if there is one), and in case of ties, invariably picks the \textit{around}-message. This speaker's strategy is a pure strategy: relative to each observation $o$, this speaker chooses a certain message with probability 1. In this case, there is no vagueness according to Lipman's definition, since the speaker does not use a mixed strategy. This is in agreement with Lipman's remark that:

\begin{quote}

``Suppose I cannot
observe height exactly. Then I should form subjective beliefs regarding which category
an individual falls into. The theorem above then says that the optimal language will
be precise about such probability distributions.''
\end{quote}

But there is vagueness according to our definition, since these are cases in which the speaker is using a word that is vague in the semantic sense. %\BS{This is what we referred to earlier as a loophole in Lipman's argument:}
{We do not define vagueness in terms of strategies, rather, we characterize it in terms of open truth-conditional content, which indirectly translates into a specific way of interpreting vague sentences}. %\BS{use of a pure strategy is less fundamental to the characterization of vagueness than the characterization we proposed in terms of the admission of open truth-conditional content.}
{Using words in non-constant ways for a given observation, which we may call \textit{making vague use of a word}, is less fundamental to vagueness than \textit{making use of a vague word}, in the sense of using a word whose truth-conditional content is partly open.}

This brings us to the discussion of a stronger meta-semantic claim by Lipman in his paper. Lipman writes: 

\begin{quote}

``There is no
meaning to the word ``tall'' aside from what people interpret it to be. If a person says
he cannot say for sure whether a particular person is ``tall'', surely this means he is not
sure how most people would categorize this person or how he would best describe this
person to others, not that he is unable to make a choice. Put differently, we use words
to communicate: there is only an answer to the question of the minimum height of a tall person if we decide to use the word in such a way. Inherently, then, it is indeed a game
theory problem.''
\end{quote}

We agree with Lipman that explaining vagueness {involves modeling} an interactive situation problem between a speaker and a listener, hence that it is a game-theoretic problem in the broad sense. Where we disagree is on the first claim that ``there is no meaning to the word ``tall'' aside from what people interpret it to be''. {Whether for ``tall'' or for ``around'', there are reasons to posit a level of semantic representation which is distinct from use. Of course, this semantic level constrains the way we use and interpret messages, but the relationship can be quite indirect.} 

\pe{The distinction between semantics and pragmatics is not endorsed here merely to resolve the puzzle raised by Lipman. We take it to be of fundamental importance to account for the fact that meanings are compositional}\bs{, and that the level at which they are compositional is, in first approximation, that of truth-conditions, not use-conditions. }
%Concretely, this implies that meanings depend on truth-conditions, and not just on use conditions. This applies to logical expressions in the first place, which are usually not seen as vague. 
\pe{Consider negation: the circumstances that make the negation of a sentence true do not coincide with the circumstances in which negation is used. For example, the truth conditions of the sentence ``Sydney is not the capital of Australia'' do not coincide in any way with the probability of uttering that sentence. More generally, in a context, the probability of uttering a sentence and the probability of uttering its negation will not even sum to 1, and are not in a simple relationship,\footnote{\pe{We are not talking here of the probability of the negation of a sentence being true, which itself depends on truth conditions.}} whereas the truth conditions of ``Sydney is not the capital of Australia" are a simple function of the truth conditions of ``Sydney is the capital of Australia". On this we suscribe to the Fregean view whereby compositionality can help to learn language, and to make inferences that would not be reliable otherwise.}

\pe{We do agree with Lipman and with a broader tradition on the fact that meaning depends in large part on use, but use is not required to fix all of the semantic facts, in that plenty of facts about meaning depend on a level of logical structure (as we argued for ``tall'', and for ``around'', and as noted before us by \citealt{fine1975vagueness} in the case of vague words, with his allusion to penumbral connections).\footnote{Penumbral connections are facts such as: ``if $x$ is tall, then if $y$ is taller than $x$, $y$ is tall''.} Along with other authors (see \citealt{brochhagen2018coevolution}), we speculate that the inclusion of a level of semantic structure may help to converge to communicative equilibria that would be much more effortful otherwise.} \bs{Consider the double desideratum that natural language meanings should (a)~be compositional and (b)~be based on a  messsage-to-meaning mapping that corresponds to an equilibrium of a signaling game, where speaker types (in the game-theoretic sense) are extremely fine-grained, since they should reflect all the parameters that enter into message choice. Such a desideratum might be very difficult to achieve in principle.  Having a level of `on-line' pragmatic reasoning on top of linguistic meaning may help provide the best of two worlds: compositionality and communicative success.}

% \BS{}{Rather, we view the meaning of an expression as a theoretical construct which has to be evaluated in terms of its explanatory role within a theory whose goal is to explain, ultimately, how language is interpreted and used. So while meaning relates to how words are interpreted, the relation can be quite indirect.} Whether for ``tall'' or for ``around'', \BS{there is}{there are reasons to posit} a level of semantic representation that constrains how people use it and interpret them, but is by itself distinct from use. 

Our account of vagueness is in this regard faithful to the Gricean division of labor between literal meaning and pragmatic meaning. As stressed by \citet[p. 453]{rothschild2013game}, in order to give a game-theoretic account of scalar implicatures (why ``some'' can mean ``some but not all''), we can't assume that messages are inherently meaningless signals that only assigned meaning in the context of an equilibrium. Instead, we need to piggyback on the logical relations between ``some'' and ``all'', and so we need a level of in-built semantic content to constrain the use of the signals. The same holds of our account of the rationality of using vague as opposed to precise words. Lipman's setting, instead, corresponds to a case of radical interpretation, where the division of labor between semantics and pragmatics is absent.

\pe{The second main difference between our account and Lipman's is that we have put a limitation on the expressiveness of the language. Because of that, it may be objected that this limitation on expressiveness is the real explanandum for a theory of vagueness. We grant this point too. In that regard, however, Lipman's results and our approach complete each other: to explain why language is vague, we need to assume that to express complete probability distributions would be inefficient or overly costly. But under that assumption, the result of the previous section tells us something of significance, which is that the existence of at least one truth-conditionally vague message is needed in order to restore some of that expressiveness and to convey fine-grained probabilistic information in an indirect way.} 

\section{Conclusion}\label{sec:conclusion}

%% Mention the sense in which hesitation in borderline cases remains true

%% Mention cases in which speaker is not uncertain, but uses vague language nonetheless: "John is tall" even when you know the height down to the last centimeter

Let us summarize our main claim here: Lipman's characterization of vagueness puts forward some features of the use of vague words that are not characteristic of vagueness, but rather coincidental. To get an adequate pragmatic and game-theoretic account of vagueness, we need a level of semantic representation, and a semantic characterization of vagueness. The characterization we adopted relies on the openness of a semantic parameter along a dimension of comparison that is taken to be represented by the speaker and the hearer in their communicative exchanges. 
%the notion of scalar vagueness, namely the openness of a parameter along a 

%dimension of comparison. 

Admittedly, words like ``around'' and ``tall'' are simple and even basic illustrations of what we have called semantic vagueness, since they manipulate essentially one open parameter along a scale comparison. But vagueness is notoriously present in a broader class of words than in adjectives like ``tall'' and prepositions like ``around'', namely in multi-dimensional vague words found in nominal, verbal and adjectival categories.\footnote{See \cite{burks1946empiricism} for a seminal presentation of the distinction between one-dimensional and multi-dimensional vagueness. See \cite{verheyen2019revealing,dambrosio2023multidimensional} for some recent discussions of multidimensionality.} For those expressions, the specification of the various dimensions that lend themselves to the truth-conditional variability we saw for ``tall'' and ``around'' can be a more delicate problem. However, we reckon the form of the problem would remain the same even in those more complex cases. The speaker, after making an observation, may be uncertain of the exact location of an object in a complex conceptual space, but may help the listener figure out that location and her uncertainty more accurately by using a term whose boundaries are open in that multi-dimensional space than by the selection of a more precise alternative.

%Consider a color adjective like ``red'': it is vague by the usual tests for vagueness

%Vagueness is also present in nominal and verbal categories, and not just in prepositions and adjectives, and the identification of the various dimensions that permit the kind of semantic variability we saw for ``tall'' and ``around'' can be a more delicate problem. 

Further issues remain to be explored. As we mentioned in our discussion of alternatives, the alternatives to a vague word like ``tall'' could be so unconstrained as to include a full verbal specification of the speaker's probability distribution. In that case, there would be no case for the optimality of using vague words. This is where Grice's maxim of Manner (``be concise'') becomes of central relevance. To model the effect of syntactic complexity on the choice of vocabulary, the present account needs an explicit representation of the cost attached to the use of distinct messages. The incorporation of cost is also needed to explain cases in which a speaker is perfectly informed (of the number of attendees, or of someone's height), but nevertheless chooses a vague expression to communicate that information ``{about} a hundred" instead of ``103''; ``very tall'' instead of ``197cm tall'').

%This is where Grice's of Manner (``be concise'') becomes of central relevance, since the alternatives to a semantically vague word ought to have, roughly, a comparable syntactic complexity, and come with an explicit account of cost, which we did not include here. 
%
This problem raises further epistemological issues, such as whether speakers can have perfectly precise access to their observational probability distributions. The rationalization we proposed in fact assumes that vague words are ways of conveying probabilistic information in a way that is not explicitly probabilistic.

\newpage

\begin{appendices}

%\color{violet}

\section{Proof of the ratio inequality for ``tall''}

The proposition expressed by `Mary is tall' is that Mary's height, which we notate $x$, exceeds the threshold for tallness, notated $t$. Both $x$ and $t$ are treated as random variables, which are independent. Therefore, 
$Pr(x=k|\text{Mary is tall})$ is equal to $Pr(x=k|x\geq t)$. Let $k_2>k_1$:

\begin{align*}
Pr(x=k_2|x\geq t) & = \sum_i Pr(x=k_2, t=i | x \geq t)\\
 & = \sum_i \dfrac{Pr(x=k_2, t= i, x \geq t)}{Pr(x\geq t)}\\
 & =\sum_i \dfrac{Pr(x=k_2, t=i, k_2 \geq i)}{Pr(x\geq t)}\\
 &= \dfrac{\sum_{i \in [0, k_2]} Pr(x=k_2)\times Pr(t=i)}{Pr(x\geq t)}, \text{and}\\
Pr(x=k_1|x\geq t) &= \dfrac{\sum_{i \in [0, k_1]} Pr(x=k_1)\times Pr(t=i)}{Pr(x\geq t)},\text{so} \\ 
\dfrac{Pr(x=k_2|x\geq t)}{Pr(x=k_1|x\geq t)}&=\dfrac{\sum_{i \in [0, k_2]} Pr(x=k_2)\times Pr(t=i)}{\sum_{i \in [0, k_1]} Pr(x=k_1)\times Pr(t=i)}\\
&=\dfrac{\sum_{i \in [0, k_2]} Pr(t=i)}{\sum_{i \in [0, k_1]} Pr(t=i)}\times \dfrac{Pr(x=k_2)}{Pr(x=k_1)}.\\
\text{However, }
\dfrac{\sum_{i \in [0, k_2]} Pr(t=i)}{\sum_{i \in [0, k_1]} Pr(t=i)}&=\dfrac{\sum_{i \in [0, k_1]} Pr(t=i)+\sum_{i \in (k_1, k_2]}Pr(t=i)}{\sum_{i \in [0, k_1]} Pr(t=i)}> 1.\\
\text{Hence: } \dfrac{Pr(x=k_2|x\geq t)}{Pr(x=k_1|x\geq t)}&>\dfrac{Pr(x=k_2)}{Pr(x=k_1)}.\\
\end{align*}

\section{Proof of the Theorem}\label{sec:appendixproof}

Let us assume that no message is vague. This means that for every $m$, %contains no open variable and therefore %for every world $w$ and pair of assignments $g_1$, $g_2$, $\sem{m}^{w,g_1}=\sem{m}^{w,g_2}$.%For every message $m$, let 
\pe{the meaning of $m$ depends merely on $w$ and not on any open parameter}, so let $\sem{m}^w$ be its truth-value in $w$. %, which is the same across all \pe{assignments of values to the variables%possible values for $i$. 
The formula for $L_0$ simplifies as follows:
\begin{equation*}
    L_0(w, o \mid m) \propto P(w, o) \cdot \sem{m}^w
\end{equation*}

\noindent Let $o_1$ and $o_2$ be two observations that have the same support, such that for every $w$, $P(w \mid o_1) > 0 \iff P(w \mid o_2) > 0$. For any observation $o$, we denote $\mathcal{S}(o)$ as the support of $o$, defined as the set $\{w : P(w \mid o) > 0\}$.

\section*{Proof by Induction}

\paragraph{Base case.} We first prove that $S_1(\cdot \mid o_1) = S_1(\cdot \mid o_2)$.

For every message $m$, 
\begin{equation*}
    U_1(m, o_1) = - \sum_{w, o} P(w, o \mid o_1) \cdot \big( \log P(w, o \mid o_1) - \log L_0(w, o \mid m) \big)
\end{equation*}
The term $P(w, o \mid o_1) \cdot \big( \log P(w, o \mid o_1) - \log L_0(w, o \mid m) \big)$ evaluates to $0$ if $o \neq o_1$, and also if $P(w \mid o_1) = 0$ (by the convention $0 \cdot \log(0) = 0$). This allows us to rewrite the sum as:
\begin{equation*}
    U_1(m, o_1) = - \sum_{w \in \mathcal{S}(o_1)} P(w \mid o_1) \cdot \big( \log P(w \mid o_1) - \log L_0(w, o_1 \mid m) \big)
\end{equation*}

\noindent  For any $w$, if $\sem{m}^w = 0$, $L_0(w, o_1 \mid m) = 0$ and $\log L_0(w, o_1 \mid m) = -\infty$. Therefore, if for some $w \in \mathcal{S}(o_1)$, $\sem{m}^w = 0$, then $U_1(m, o_1) = -\infty$. This captures the maxim of Quality: a message $m$ provides infinitely negative utility if it is not entailed by the support of the observation. In such a case, we obviously also have $U_1(m, o_2) = -\infty$, since $o_1$ and $o_2$ have the same support, and so $m$ is equally unusable with $o_1$ and with $o_2$.\\
\\
Assume now that for every $w \in \mathcal{S}(o_1)$, $\sem{m}^w = 1$. Then for every such $w$, we have:
\begin{equation*}
    L_0(w, o_1 \mid m) = \frac{P(w, o_1)}{\sum\limits_{w', o'} P(w', o') \cdot \sem{m}^{w'}}
\end{equation*}
Let the denominator of the fraction be denoted as $D(m)$ -- it is simply the probability of $m$ being true. Substituting this, we get:
\begin{equation*}
    U_1(m, o_1) = - \sum_{w \in \mathcal{S}(o_1)} P(w \mid o_1) \cdot \big( \log P(w \mid o_1) - \log P(w, o_1) + \log D(m) \big)
\end{equation*}

Given that $ \log P(w \mid o_1) = \log P(w, o_1) - \log P(o_1)$, this itself simplifies to:

\begin{align*}
    U_1(m, o_1) = \sum_{w \in \mathcal{S}(o_1)} P(w \mid o_1) \cdot \big( \log P(o_1) -\log D(m) \big)\\ = ( \log P(o_1) -\log D(m) \big) \cdot \sum_{w \in \mathcal{S}(o_1)} P(w \mid o_1)\\
    = \log P(o_1) - \log D(m)
\end{align*}
Hence: 

\begin{equation*}
    U_1(m, o_1) - U_1(m, o_2) = \log P(o_1) - \log P(o_2) 
\end{equation*}
This difference does not depend on $m$. Denote it as $K(o_1, o_2)$. Thus, for any two observations $o_1$, $o_2$ with the same support, and for any message $m$, we have:
\begin{equation*}
    U_1(m, o_1) = U_1(m, o_2) + K(o_1, o_2)
\end{equation*}
As a result, the set of messages $m$ that maximize $U_1(m, o_1)$ is exactly the same as the set of messages $m$ that maximize $U_1(m, o_2)$. Therefore:
\begin{equation*}
    \operatorname*{argmax}_{m'} U_1(m', o_1) = \operatorname*{argmax}_{m'} U_1(m', o_2)
\end{equation*}
Hence, for any message $m$,
\begin{equation*}
    S_1(m \mid o_1) = S_1(m \mid o_2)
\end{equation*}

\paragraph{Inductive Step}. We assume that for some $k$, $S_k(\cdot \mid o_1) =S_k(\cdot \mid o_2)$. We need to prove that $S_{k+1}(\cdot \mid o_1) =S_{k+1}(\cdot \mid o_2)$

First, if $S_k(m)=0$ (i.e. $m$ is a surprise message at level $k$), then $L_k(\cdot \mid m) = L_0(\cdot \mid m)$, and therefore $U_{k+1}(m,o_1) = U_1(m,o_1)$ and $U_{k+1}(m,o_2) = U_1(m,o_2)$, hence $U_{k+1}(m, o_1) - U_{k+1}(m, o_2) = \log P(o_1) - \log P(o_2)$\\
\\
Otherwise:

\begin{equation*}
    U_{k+1}(m,o_1) = - \sum_{w,o} P(w,o \mid o_1) \cdot \big( \log P(w,o \mid o_1) - \log L_k(w, o \mid m) \big)
\end{equation*}
The term $P(w, o \mid o_1) \cdot \big( \log P(w, o \mid o_1) - \log L_k(w, o \mid m) \big)$ evaluates to $0$ if $o \neq o_1$, and also if $P(w \mid o_1) = 0$ (by the convention $0 \cdot \log(0) = 0$). This allows us to rewrite the sum as:

\begin{equation*}
    U_{k+1}(m, o_1) = - \sum_{w \in \mathcal{S}(o_1)} P(w \mid o_1) \cdot \big( \log P(w \mid o_1) - \log L_k(w, o_1 \mid m) \big)
\end{equation*}
Given the definition of $L_k$, this yields:

\begin{equation*}
    U_{k+1}(m, o_1) = - \sum_{w \in \mathcal{S}(o_1)} P(w \mid o_1) \cdot \big( \log P(w \mid o_1) - \log \dfrac{P(w,o_1)\cdot S_k(m \mid o_1)}{\sum\limits_{w',o'}P(w',o')\cdot S_k(m \mid o')} \big)
\end{equation*}
With $\alpha(m) =\sum\limits_{w',o'}P(w',o')\cdot S_k(m \mid o')$, we have

\begin{align*}
    U_{k+1}(m, o_1) &= - \sum_{w \in \mathcal{S}(o_1)} P(w \mid o_1) \cdot \big( \log P(w \mid o_1) - \log P(w,o_1) + \log S_k(m \mid o_1) - \log \alpha(m) \big)\\
    &= \sum_{w \in \mathcal{S}(o_1)} P(w \mid o_1) \cdot ( \log P(o_1) - \log S_k(m \mid o_1) + \log \alpha(m))\\
   & = ( \log P(o_1) - \log S_k(m \mid o_1) + \log \alpha(m)) \cdot \sum_{w \in \mathcal{S}(o_1)} P(w \mid o_1)\\ &= \log P(o_1) - \log S_k(m \mid o_1) + \log \alpha(m)
\end{align*}
By the induction hypothesis, $S_k(m \mid o_1) = S_k(m \mid o_2)$. Therefore, either $S_k(m~\mid~o_1) = S_k(m~\mid~o_2) = 0$, in which case $U_{k+1}(m, o_1) = U_{k+1}(m, o_2) = -\infty$ and $S_{k+1}(m,o_1) = S_{k+1}(m,o_2) = 0$, or

\begin{equation*}
    U_{k+1}(m,o_1) - U_{k+1}(m,o_2) = \log P(o_1) - \log P(o_2)
\end{equation*}
Therefore, unless $U_{k+1}(m, o_1) = U_{k+1}(m, o_2) = -\infty$, both if $m$ is a surprise message at level $k$ and if not, we have $U_{k+1}(m,o_1) = U_{k+1}(m,o_2) + K(o_1,o_2)$, with $K(o_1,o_2)= \log P(o_1) - \log P(o_2)$. Hence  the set of messages $m$ that maximize $U_{k+1}(m, o_1)$ is exactly the same as the set of messages $m$ that maximize $U_{k+1}(m, o_2)$. Therefore:
\begin{equation*}
    \operatorname*{argmax}_{m'} U_{k+1}(m', o_1) = \operatorname*{argmax}_{m'} U_{k+1}(m', o_2)
\end{equation*}
Hence, for any message $m$,
\begin{equation*}
    S_{k+1}(m \mid o_1) = S_{k+1}(m \mid o_2)
\end{equation*}

\hfill QED.

\section{An example of an IRB equilibrium for a language with a vague message}\label{sec:appendixexample}

We provide a fully specified example of a game which contains a vague message and where, at equilibrium, the speaker's choice of a message can depend not just on the support of her probability distribution but also on its specific shape.

The variable of interest (corresponding to the random variable $w$) ranges from $0$ to $8$. The speaker can make four observations. The observation notated \texttt{p\_1\_7} is defined as giving rise to a posterior distribution over $w$ which is \emph{peaked} and has support $[1, 7]$ (so excludes $0$ and $8$). The observation notated \texttt{u\_1\_7} is one that leads to a uniform distribution on $[1,7]$ (it also excludes $0$ and $8$). The observation \texttt{u\_0\_8} gives rise to a uniform posterior distribution over the full range. Finally, the distribution \texttt{ap\_0\_8} gives rise to a posterior distribution with support $[0,8]$ which is \emph{anti-peaked}, i.e., assigns higher probabilities to values further away from the center (which is the number $4$).

The precise conditional distributions (the distributions induced by each observation) are given in the following table, together with the probability of each observation (last column):

\begin{center}
\renewcommand{\arraystretch}{1.3}
\begin{tabular}{p{2cm}|ccccccccc|c}
\toprule
\textbf{{\footnotesize Observation}} & \textbf{0} & \textbf{1} & \textbf{2} & \textbf{3} & \textbf{4} & \textbf{5} & \textbf{6} & \textbf{7} & \textbf{8} & \textbf{P(obs)} \\
\midrule
\texttt{p\_1\_7} & 0 & 0.01 & 0.01 & 0.16 & 0.64 & 0.16 & 0.01 & 0.01 & 0 & $\nicefrac{5}{54}$ \\
\texttt{u\_1\_7} & 0 & $\nicefrac{1}{7}$ & $\nicefrac{1}{7}$ & $\nicefrac{1}{7}$ & $\nicefrac{1}{7}$ & $\nicefrac{1}{7}$ & $\nicefrac{1}{7}$ & $\nicefrac{1}{7}$ & 0 & $\nicefrac{7}{270}$ \\
\texttt{u\_0\_8} & $\nicefrac{1}{9}$ & $\nicefrac{1}{9}$ & $\nicefrac{1}{9}$ & $\nicefrac{1}{9}$ & $\nicefrac{1}{9}$ & $\nicefrac{1}{9}$ & $\nicefrac{1}{9}$ & $\nicefrac{1}{9}$ & $\nicefrac{1}{9}$ & $\nicefrac{5}{12}$ \\
\texttt{ap\_0\_8} & $\nicefrac{35}{251}$ & $\nicefrac{65}{502}$ & $\nicefrac{65}{502}$ & $\nicefrac{25}{251}$ & $\nicefrac{1}{251}$ & $\nicefrac{25}{251}$ & $\nicefrac{65}{502}$ & $\nicefrac{65}{502}$ & $\nicefrac{35}{251}$ & $\nicefrac{251}{540}$ \\
\bottomrule
\end{tabular}
\end{center}

\bigskip

\noindent For example, we have $P(w=3 \mid o = \texttt{p\_1\_7}) = 0.16$.\\
\\
The prior joint probability of any pair of world $w$ and observation $o$ is given by:
\[
P(w,o) = P(w \mid o) \times P(o).
\]
For instance, $P(w=3, o=\texttt{p\_1\_7}) = 0.16 \times \nicefrac{7}{270}$.\\
\\
The resulting prior joint probability distribution is such that, by design, the prior probabilities of each value for $w$, obtained by marginalizing over $o$, is uniform on $[0,8]$. It follows that the observation \texttt{u\_0\_8} can be thought of as a totally uninformative observation, resulting in a posterior distribution over $w$ that is identical to the prior distribution.

\medskip

\noindent There are three messages:
\begin{itemize}
  \item the \emph{null} message, which is true in every world;
  \item the \emph{‘between 1 and 7’} message, which is true if $w \in [1,7]$, false otherwise;
  \item the \emph{‘around 4’} message, which is a vague message and whose interpretation is relativized to an open parameter $t$, viewed as a random variable that takes its value in $[0,4]$. If $t=i$, then \emph{‘around 4’} is  true if $w$ is in the interval $[4{-}i, 4{+}i]$. The prior probability distribution over $t$ is uniform over $[0,4]$.

\end{itemize}

\medskip

\noindent In this setting, there is an IRB equilibrium, and it is reached as early as $(S_1, L_1)$. In this equilibrium:
\begin{itemize}
  \item the speaker uses the null message if she observed \texttt{u\_0\_8} or \texttt{ap\_0\_8};
  \item she uses the \emph{between} message if she observed \texttt{u\_1\_7};
  \item she uses the \emph{around} message if she observed \texttt{p\_1\_7}.
\end{itemize}

\noindent Correspondingly, the listener:
\begin{itemize}
  \item infers from the \emph{between} message that the speaker observed \texttt{u\_1\_7};
  \item infers from the \emph{around} message that the speaker observed \texttt{p\_1\_7};
  \item infers from the null message that the speaker either observed \texttt{u\_0\_8} or \texttt{ap\_0\_8}.
\end{itemize}

\noindent This therefore provides an example where the speaker chooses a different message for \texttt{u\_1\_7} and for \texttt{p\_1\_7}, even though these two observations give rise to posterior distributions over $w$ which have the same support.

\end{appendices}

\newpage

\bibliographystyle{apalike}
\bibliography{egre_biblio}

\end{document}